\documentclass[preprint,12pt]{elsarticle}

\usepackage{lineno,hyperref}
\usepackage{booktabs}
\usepackage{graphicx}
\usepackage{adjustbox}
\usepackage{multirow}
\usepackage{amsmath}
\usepackage{amssymb}
\newcommand{\R}{\mathbb{R}}
\newcommand{\E}{\mathbb{E}}
\newcommand{\sign}{\text{sign}}
\DeclareMathOperator*{\argmax}{arg\,max}
\usepackage[ruled,vlined,linesnumbered,commentsnumbered]{algorithm2e}
\usepackage{bm}
\usepackage{float}
\usepackage{longtable}
\restylefloat{table}

\journal{Swarm and Evolutionary Computation}

\begin{document}

\begin{frontmatter}

\title{Brain Programming is Immune to Adversarial Attacks: Towards Accurate and Robust Image Classification using Symbolic Learning}

\author[1]{Gerardo Ibarra-Vazquez}
\author[2]{Gustavo Olague}
\cortext[mycorrespondingauthor]{Corresponding author}
\ead{olague@cicese.mx}
\author[2]{Mariana Chan-Ley}
\author[1]{Cesar Puente}
\author[1]{Carlos Soubervielle-Montalvo}

\address[1]{Universidad Autónoma de San Luis Potosí, Facultad de Ingeniería. Dr. Manuel Nava 8, Col. Zona Universitaria Poniente, 78290, San Luis Potosí, S.L.P., México}
\address[2]{EvoVisión Laboratory, CICESE Research Center. Carretera Ensenada-Tijuana 3918, Zona Playitas, 22860, Ensenada, B.C. México}

\begin{abstract}
In recent years, the security concerns about the vulnerability of Deep Convolutional Neural Networks (DCNN) to Adversarial Attacks (AA) in the form of small modifications to the input image almost invisible to human vision make their predictions untrustworthy. Therefore, it is necessary to provide robustness to adversarial examples in addition to an accurate score when developing a new classifier. In this work, we perform a comparative study of the effects of AA on the complex problem of art media categorization, which involves a sophisticated analysis of features to classify a fine collection of artworks. We tested a prevailing bag of visual words approach from computer vision, four state-of-the-art DCNN models (AlexNet, VGG, ResNet, ResNet101), and the Brain Programming (BP) algorithm. In this study, we analyze the algorithms' performance using accuracy. Besides, we use the accuracy ratio between adversarial examples and clean images to measure robustness. Moreover, we propose a statistical analysis of each classifier's predictions' confidence to corroborate the results. We confirm that BP predictions' change was below 2\% using adversarial examples computed with the fast gradient sign method. Also, considering the multiple pixel attack, BP obtained four out of seven classes without changes and the rest with a maximum error of 4\% in the predictions. Finally, BP also gets four categories using adversarial patches without changes and for the remaining three classes with a variation of 1\%. Additionally, the statistical analysis showed that the predictions' confidence of BP were not significantly different for each pair of clean and perturbed images in every experiment. These results prove BP's robustness against adversarial examples compared to DCNN and handcrafted features methods, whose performance on the art media classification was compromised with the proposed perturbations. We also ratify the competitive score of BP against the state-of-the-art classifiers for the art media categorization problem.
\end{abstract}


\end{frontmatter}


\section{Introduction}
\label{sec:Intro}

Image classification is an active research area in artificial intelligence, whose primary goal is to analyze contextual information or visual content of an image and assign it to the class or category to which it belongs \cite{aima2020}. There have been significant efforts on areas such as Computer Vision (CV), Machine Learning (ML), Evolutionary computation (EC), and Swarm Intelligence (SI) to tackle this problem \cite{szeliski2020computer,olague2016evolutionary,Darwish2019ASO}. Two predominant methods have been among the most popular and successful approaches for solving image classification problems: 1) Bag of Visual words (BoV) from CV and 2) Deep Convolutional Neural Networks (DCNN) also known as Deep Learning (DL), a subdivision of ML \cite{Druzhkov2016ASO,Zhao2017ASO}. Nonetheless, EC and SI have mostly contributed in two manners: 1) optimizing feature selection and 2) optimizing DCNN architectures.

In this manner, Genetic Programming (GP) has been one of EC's principal tools to optimize the selection of features and automatically extract the best characteristics to approach image classification tasks. For example, In 2018, authors of \cite{Bi2018GeneticPF} propose a GP method to achieve simultaneously global and local feature extraction for image classification using the JAFFE (1998), YALE (1997), FLOWER (2007), and TEXTURE (2006) datasets. As we could appreciate, all datasets are outdated since nowadays; none are used to test algorithms. Moreover, their approach is compared to traditional hand-engineered features from CV like SIFT (Scale-Invariant Feature Transform), which is an image processing technique that follows the local feature paradigm, and it does not behave-scale well for problems like image categorization since different images with multiple attributes represent an object category. The solution demands a consensus of distinct characteristics in the form of a set of features. Therefore, this research work lacks more recent databases and a comparison with current classification algorithms. In 2019, the article \cite{Price2019GOOFeDEA} proposes a GP approach to automatically generate discriminative rich features for image classification using the MIT urban and nature scene datasets (2003). These image databases are also outdated; hence, the comparison is made with traditional CV classification methods like Histogram of Oriented Gradients (HOG) and Support Vector Machine (SVM), similarly to the previous work.   

In 2019, the research work \cite{Iqbal2019GeneticPW} proposed a method for employing transfer learning in GP to extract and transfer knowledge to classify complex texture images. The proposed methodology uses the following texture datasets Kylberg (2011), Brodatz (1999), and Outex (2002), and all images are resized to $115 \times 115$ pixels to perform their experiments to avoid the computational cost and simplify the problem. In 2020, the article \cite{Bi2020AnEF} proposes a GP-based feature learning approach to select and combine five methods automatically: Hist (Histogram features), DIF (Domain-Independent Features), SIFT, HOG, and LBP (Local Binary Patterns). The technique generates a compound solution that extracts high-level features to classify images from classical problems with low-resolution datasets--about $100 \times 100$ pixels up to $200 \times 200$ pixels. Authors compared their approach with other GP-based methods and DL methods like LeNet-5 (a CNN model with an input of grayscale images of $32 \times 32$ pixels, toy-method in comparison with the state-of-the-art) and two hand-craft CNNs models of five- and eight-layers without providing the network parameters' information. Hence, it is not easy to judge the performance.

In addition to optimizing feature selection, EC and SI have developed strategies to search for meaningful DCNN architectures for image classification \cite{Darwish2019ASO}. Nevertheless, recent approaches, which are summarized in \cite{Nakane2020ApplicationOE}, explore hybridization of swarm and evolutionary computation algorithms by aggregating hyper-parameters' optimization during training. To give an example, in 2019, authors from \cite{8712430} proposed a novel method named evoCNN, which uses genetic algorithms for evolving DCNN architectures and connection values to address image classification problems. Their experiments were based on nine datasets that use grayscale images of $28 \times 28$ pixels: MNIST, MNIST-RD, MNIST-RB, MNIST-BI, MNIST-RD + BI, Rectangles, Rectangles-I, Convex, and MNIST-Fashion. However, in 2019, authors from \cite{Junior2019ParticleSO} proposed a novel algorithm based on particle swarm optimization (PSO) named psoCNN, capable of automatically searching DCNN architectures for image classification with fast convergence when compared with others evolutionary approaches like evoCNN, IPPSO, among others. Their experiment used the same nine datasets mentioned above.  

Despite the effort and interest made by the EC and SI communities to tackle the problem of image classification, they still are dealing with outdated problems using classical datasets while making comparisons against obsolete DCNN models. EC and SI have fallen short to be on par with DCNN models with minor works that do not exceed hand-craft DCNN architectures.

Nonetheless, a Deep Genetic Programming Methodology called Brain Programming, inspired by neuroscience knowledge that uses symbolic representations and incorporates rules from expert systems with a hierarchical structure inspired by the human visual cortex was developed by the EvoVision research team. In 2016, Evovision started evolving an Artificial Visual Cortex (AVC) for image classification and object detection. Hernández et al. used natural images of medium size (VGA) using GRAZ-01 (2003), and GRAZ-02 (2004) datasets, which are the base for the Visual Object Challenge (VOC challenge)--both still relevant in CV literature--\cite{HERNANDEZ2016216}. The results were compared with several feature extraction methods Basic Moments (2006), Hierarchical MAX - Genetic Algorithm--HMAX-GA (2012), Enhanced Biologically Inspired Model--EBIM (2011), SIFT (2006),  Similarity Measure Segmentation--SM (2006), and Moment Invariants (2006) most from CV and one including EC.  In 2017, Hernández et al. \cite{Hernndez2017CUDAbasedPO} implemented a CUDA version of BP to speed up the original system's processing time. The experiment analyzed the performance using different image sizes, which started with $256 \times 256$ pixels, doubling the sizes to up to $4096 \times 4096$ pixels, demonstrating the possibility of real-time functionality as well as the application to high-definition images. Additionally, the method was compared in time performance with a CUDA implementation of HMAX and CUDA version of a CNN with outstanding results.  

In 2019, the article \cite{Olague2019} proposes a random search to find best-fit parameters for the AVC in image classification. The experiment found great individuals to classify GRAZ-01, GRAZ-02, and Caltech-101 (2004) datasets. GRAZ datasets have image sizes of $640 \times 480$ pixels, and Caltech-101 has images of $300 \times 200$ pixels. Note that GRAZ images present a significant challenge due to the short object occurrence in the whole image, becoming challenging to resize images for processing. In contrast, Caltech-101 presents a truly image recognition dataset. In 2020, BP was proposed as a technique to approach the complex problem of Art Media Categorization (AMC) \cite{ChanLey2020CategorizationOD}. The experiment consists of classifying high-resolution art datasets such as WikiArt (2016) and Kaggle Art Images (2018). Moreover, BP results were compared with a renowned DCNN model named AlexNet, obtaining a competitive outcome. Also, BP has been evaluated on real-world problems of object tracking using standard datasets and algorithms--FRAGtrack and MILtrack--while also achieving outstanding results in real-working conditions compared to the method of Regions with Convolutional Neural Networks (R-CNN) \cite{Olague2018,Olague2019b}.

However, despite the progress made to build better image classifiers, a research opportunity that has not been considered in EC and SI is the classifier's predictions' robustness. Nowadays, there is a big concern about the performance of DCNN, which has opened a new research area in charge of dealing with Adversarial Attacks (AA) that intentionally create small perturbations in the input image to mislead the model to predict wrongly \cite{akhtar2018threat,Li_2018,OZDAG2018152,Chen2019,Xu2020,REN2020346}. Some of these perturbations are imperceptible to human vision and can completely change the DCNN's prediction to drop its performance. They are generated through a variety of forms, including making small modifications to the input pixels, using spatial transformations, among others. In addition to the analysis of DCNN vulnerabilities, there have been immense efforts to develop defense mechanisms to mitigate AA. Still, the perturbations have become more complex and highly efficient in fooling DCNN. Therefore, in this paper, we want to contrast the classification models between performance and robustness to perturbations to guarantee predictions' trustworthiness while not focusing only on accuracy.

\subsection{Problem Statement}
\label{sec:ProblemStatement}

In this section, we detail the serious problem in the DCNN structure to AA. First, given an input image $\mathbf{x}$ in an input subspace $\mathbf{X}$ such that $\mathbf{x} \in \mathbf{X}$ and its corresponding label $y$, DCNN model establishes a relationship within the data using the following equation:

\begin{equation}
    y=f(\mathbf{x})=\mathbf{w}^\intercal \mathbf{x} \;\;\;,
\end{equation}

where function $f()$ is the DCNN model, whose associated weights parameters are $\mathbf{w}$. However, an erroneous behavior is notable when the input image suffers a small change in its pixels $\mathbf{x}_{\mathbf{\rho}} = \mathbf{x} + \rho$ such that:

\begin{equation} \label{eq:aa}
\begin{split}
    f(\mathbf{x}) \neq f(\mathbf{x}_{\mathbf{\rho}}) & \;\;\;s.t.\\
    ||\mathbf{x} - \mathbf{x}_{\mathbf{\rho}}||_{p}<\alpha &
\end{split}
\end{equation}

where $p\in N ,p\geq 1, \alpha \in R, \alpha\geq 0$. So, it can be defined an Adversarial Example (AE) as an intentional modified input $\mathbf{x}_{\mathbf{\rho}}$ that is classified differently than $\mathbf{x}$ by the DCNN model, with a limited level of change in the pixels of $||\mathbf{x} - \mathbf{x}_{\mathbf{\rho}}||_{p}<\alpha$, so that it may be imperceptible to a human eye. 

The simplest explanation of how AEs work to attack a DCNN is that most digital images use 8‑bit per channel per pixel. So, each step of 1/255 limits the data representation; the information in between is not used. Therefore, if every element of a perturbation $\rho$ is smaller than the data resolution, it is coherent for the linear model to predict distinct given an input $\mathbf{x}$ than to an adversarial input $\mathbf{x}_{\mathbf{\rho}} = \mathbf{x} + \rho$. We assume that forasmuch as $||\mathbf{\rho}||_{\infty} <\alpha$, where $\alpha$ is too small to be discarded, the classifiers should predict the same class to $\mathbf{x}$ and $\mathbf{x}_{\mathbf{\rho}}$.

Nonetheless, after applying the weight matrix $\mathbf{w} \in \mathbf{R}^{M \times N}$ to the AE, we obtain the dot product defined by $\mathbf{w}^\intercal \mathbf{x}_{\mathbf{\rho}}= \mathbf{w}^\intercal\mathbf{x} + \mathbf{w}^\intercal\mathbf{\rho}$. Hence, the AE will grow the activation by $\mathbf{w}^\intercal\mathbf{\rho}$. Note that the dimensionality of the problem does not grow with $||\mathbf{\rho}||_{\infty}$; thus, the activation change caused by perturbation $\mathbf{\rho}$ can grow linearly with $n$. As a result, the perturbation can make many imperceptible changes to the input to obtain big output changes.

DCNN behavior is hugely linear to be immune to AEs, and nonlinear models such as sigmoid networks are set up to be in the non-saturating most of the time, becoming them more like a linear model. Hence, every perturbation as accessible or challenging to compute should also affect the DCNN. Therefore, when a model is affected by an AE, this image often affects another model, whether the two models have different architectures or were trained with other databases. They only have to be set up for the same task to change the result \cite{Goodfellow2015ExplainingAH}.

In this manner, the AE generation finds an input $\mathbf{x}_{\mathbf{\rho}}$ in the input subspace $\mathbf{X}'$ such that $\mathbf{x}_{\mathbf{\rho}}\in \mathbf{X}'$ and $f(\mathbf{x}) \neq f(\mathbf{x}_{\mathbf{\rho}})$. Nevertheless, we denote robustness in terms of function continuity. Given a model's function $f()$ in an input subspace $\mathbf{X}$ such that $\mathbf{x} \in \mathbf{X}$, if $\mathbf{x}_k \rightarrow \mathbf{x}$ implies $f(\mathbf{x}_k) \rightarrow f(\mathbf{x})$. Equivalently, $f(\mathbf{x})$ is robust at $\mathbf{x}$, for all $\mathbf{y} \in \mathbf{X}$, if given a $\kappa > 0$, there is a $\mu > 0$ such that $||\mathbf{y}-\mathbf{x}|| < \mu$ implies $|f(\mathbf{y}) - f(\mathbf{x})| < \kappa$. Hence, if $f(\mathbf{x})$ is robust for every $\mathbf{x}$, then $f(\mathbf{x})$ is said to be robust on $\mathbf{X}$.

Therefore, the procedure to measure robustness is by using appropriate statistical tests depending on the results' properties, standard performance measures, and the ratio of accuracies. Statistical tests allow us to determine whether the results obtained are significantly different and improve the knowledge of certain aspects over the existing algorithm measures: effectiveness, efficiency, accuracy, or reliability when using artificial neural networks, SVM, or other metaheuristics \cite{LUENGO20097798}.

\subsection{Research Contributions}
This paper provides insight into adversarial attacks and the motivation to analyze image classification models' robustness. Therefore, we extend the first results reported at the International Symposium on Visual Computing (ISVC'20), on which we explore the robustness through the complex image classification task of the AMC \cite{Olague2020}. In this work, we test a prevailing BoV approach from CV, four state-of-the-art DCNN models (AlexNet, VGG, ResNet, ResNet101), and the BP algorithm using three AAs (Fast Gradient Sign Method--FGSM, multiple pixel attack, and adversarial patch). We remark the following contributions:

\begin{itemize}
    \item The first contribution consists of a proposal of BP's robustness as a secure mechanism to deal with AAs for AMC. Since this problem is difficult, with many artifacts across all studied classes, the results can be extended to other challenging classification tasks.
    \item The second is a comparative study of predominant image classification methodologies performances from three-different research areas (CV, ML, and EC) considering AAs.
    \item The third is a statistical analysis of the proposed image classifiers from the standpoint of robustness to AAs.
\end{itemize}

We organize the present work as follows. Section \ref{sec:Related} presents relevant research for the AMC. It covers the complexity of the problem and how it has been tackle, from handcrafted features to DCNN models and GP methods. It also presents the concerns about AAs and how the robustness predictions have not been studied on the AMC. Section \ref{sec:Methodology} outlines each classification method's structure and the AAs used in this work. Section \ref{sec:Experiments} gives details about the experimental setup, including the dataset, its construction, the employed evaluation metrics, and the experimental results' explanation. Finally, Section \ref{sec:Conclusion} presents the conclusions of this work.

\section{Related works}
\label{sec:Related}

The AMC problem in CV has arisen from the need to have automatic systems for identifying valuable artwork pieces to have a trustworthy analysis of complex features that can not be subjective as humans are prone to be. For example, classifying fine art pieces involves a sophisticated selection of features that distinguish each medium, which is extremely difficult \cite{FALOMIR201883}. Usually, an art expert analyzes the style, genre, and media from artwork to identify the artist and detect forgeries \cite{Kong2017,Elgammal2018TheSO,ChanLey2020CategorizationOD}. Therefore, the development of automated systems that provide such tasks makes an accurate and robust analysis a critical issue in terms of security. In this manner, to make a robust analysis of art media, high-resolution images are mandatory to provide enough information to maximize carefulness based on the artwork details. The art style, usually associated with the author's school, describes the artists' distinctive artifacts, visual elements, techniques, and methods. The form is related to the localization of features at different levels. The classical hierarchy of genres ranks history-painting and portrait as high, while landscapes and still-life are classified as low because they did not contain persons.

AMC has been tackled from 3 perspectives: 1) handcrafted feature extraction, 2) deep convolutional neural networks, and 3) genetic programming methodologies. First, handcrafted engineered features were the principal method to develop formulas that can extract features to obtain an image representation to classify an image easily.

One of the first works that employ handcrafted features was \cite{keren2002painter}; here, the authors proposed a Discrete Cosine Transform (DCT) coefficients scheme used for feature extraction painter identification by classifying the artist's style. They build a custom database of approximately 300 grayscale images from five painters (Rembrandt, Van-Gogh, Picasso, Magritte, and Dali) to experiment. Li and Wang \cite{li2004studying} proposed using a two-dimensional multi-resolution hidden Markov model to analyze brush strokes to provide reliable information to distinguish artists from ancient Chinese paintings. Their database consists of 276 grayscale images from five Chinese artists at a resolution of $3000 \times 2000$ pixels but scaled to 512 on the shorter dimension, maintaining the aspect ratio. Authors in \cite{arora2012towards} present a comparative study of different classification methodologies based on handcrafted engineered features. They contrasted semantic-level features with an SVM, color SIFT and opponent SIFT with BoV, and latent Dirichlet allocation with a generative BoV topic model for fine-art genre classification. In their study, a database of seven categories of paintings (Abstract, Baroque, Renaissance, Popart, Expressionism, Impressionism, and Cubism) was used from the Artchive fine-art dataset using 70 images from each class. Recently, Rosado \cite{Rosado2019ComputerVM} employed a BoV implemented using a dense-SIFT method for feature extraction and Probabilistic Latent Semantic Analysis (PLSA) to make an image analysis of 434 digitized images from paintings, drawings, books, and engravings by Antoni Tàpies. In general, we note that using handcrafted engineered features makes it possible to obtain encouraging but not perfect results. Over time, the complexity of these characteristics started to become more challenging to design. In addition to the designing process of features, the learning algorithm development was a completely independent research area needed to match the feature extraction.

DCNN have been a breakthrough in many areas of image processing, and recent works on AMC have presented approaches based on the state-of-the-art DCNN architectures. Authors in \cite{Karayev2014RecognizingIS} introduced the use of deep convolutional activation features from a DCNN model trained for object recognition to recognize the style. These features achieve high performance identifying styles in painting images and outperform most handcrafted engineered features. Bar et al. \cite{bar2014classification} proposed a compact binary representation combining the PiCoDes descriptors and the deep convolutional activation features from a DCNN model to identify artistic styles in paintings showing exceptional results to classify artwork images from WikiArt using 27 classes. Noord et al. \cite{Noord2015TowardDO} employed an adaptation of AlexNet to classify artwork styles from Rijks Museum images. They could visualize the regions with a heatmap from the artwork that impacts the prediction of style. Cetinic and Grgic \cite{Cetinic2016GenreCO} utilized the features extracted from VGG to classify WikiArt database images into seven genre classes such as portrait, landscape, city, still life, nude, flower, and animal. They outperform handcrafted engineered features such as SIFT, gist descriptor, HOG, Gray Level Co-occurrence Matrix (GLCM), and HSV color histograms with their classification method. Seguin et al. \cite{Seguin2016VisualLR} propose to extract from VGG similar components shared by various artworks named visual link. These links try to find similitude from the paintings of the same creators or the same schools. The experiment used images from the Web Gallery of Art database reporting that their method achieves better performance than handcrafted engineered features such as SIFT.

Sun et al. \cite{Sun2017ConvolutionNN} employed AlexNet and VGG to construct a structure with two pathways to obtain object and texture features. The DCNN performs the object computation, and the texture pathway uses the Gram matrices of intermediate features. Authors used in their experiments WikiPaintings, Flickr Style, and AVA Style databases. Elgammal et al. \cite{Elgammal2018PicassoMO} proposed an analysis of strokes in line drawings using a database of 300 digitized drawings with over 80 thousand strokes. They employ handcrafted engineered features, deep learned features, and the combination of both to discriminate between artists at the stroke-level with high accuracy. Also, their work serves to discover forgeries made by artists. Cetinic et al. \cite{Cetinic2018FinetuningCN} performed an extensive CNN fine-tuning experiment using five Caffe models (CaffeNet, Hybrid-CNN network, MemNet network, Sentiment network, and Flickr network) for five different art-related classification tasks (artist, genre, style, time period, and association with a specific national artistic context) on three large fine art datasets (WikiArt, Web Gallery of Art, and TICC Printmaking Dataset). In \cite{yang2020classification}, authors employed pre-train DCNN models (AlexNet, VGG, GoogLeNet, ResNet, DenseNet) to recognize basic artistic media from artworks. They collected about 1000 artwork images per class (oil-paint brush, pastel, pencil, and watercolor) through various search engines and websites to classify them. They obtained comparable results with that of trained humans.

Finally, a GP-like methodology called brain programming has obtained competitive results compared to a DCNN model for the AMC task \cite{ChanLey2020CategorizationOD}. This technique aims to emulate the brain's behavior based on neuroscience learning processes with new symbolic learning. In the experiments, two renowned databases of high-resolution artwork images are used (art database from Kaggle and WikiArt) to classify five art media classes (drawings, engraving, painting, iconography, and sculpture). The proposed technique achieves comparable results to AlexNet on a binary classification problem. 

Although DCNN have obtained exemplary results in solving a wide variety of computer vision tasks, small perturbations named adversarial attacks made on the input image turn the learning model's decision to change its prediction completely. These perturbations are generated in several forms that include small modifications to the input pixels and using spatial transformations, among others. These attacks' primary purpose is to fool the DL models prediction intentionally and remain unnoticed to human perception. Szegedy et al. \cite{Szegedy2014IntriguingPO} were the first who discovered an unusual weakness where small perturbations almost invisible to the human vision on the input pixels can fool a CNN. These attacks also reported high confidence in the model's wrong prediction, and even worse, multiple networks were affected using the same perturbed image. Later, they found that CNN's robustness against AA could be improved using these images in the training phase. However, recent studies have highlighted the lack of robustness in well-trained DCNNs \cite{Zheng2016,Su2018}. Goodfellow et al. \cite{Goodfellow2015ExplainingAH} designed a method named Fast Gradient Sign Method (FGSM), which enables efficient computing perturbations for a given image. Another threat consists of an extreme and straightforward attack proposed by Su et al. \cite{su2019one}, which consists of modifying one pixel in the image, can fool a CNN. A drawback, however, is that it only works for icon images. They successfully attacked three different network models under this strategy with high confidence. Moosavi-Dezfooli et al. \cite{moosavi2017universal} discovered singular perturbations that can misclassify any image; they called it universal perturbations. In this way, Brown et al. \cite{brown2017adversarial} proposed a method to create universal, robust, targeted adversarial image patches. These patches are so compact that they can be printed and used in real-world scenes to fool a CNN.

Despite significant efforts in making defense methods against AAs, the research works have focused on modifying its training process or modifying the input image during testing \cite{Goodfellow2015ExplainingAH,song2018pixeldefend,Baek2019AdversarialLW}, also on changing the structure of the networks \cite{Gu2015TowardsDN,ross2018improving,papernot2016distillation} or through external models to classify unseen examples \cite{meng2017magnet,akhtar2018defense}. Zhang et al. \cite{zhang2019thelimita} discussed the limitation of the adversarial training because the attacks have become more and more challenging with high efficiency on the damage.

AMC is a complex problem to solve. Its solution involves a complicated analysis of features and demands accurate and robust decisions, mostly when curators work with highly valuable art pieces. The performance of handcrafted engineered features methods has been limited to compete with DCNN through their inability to extract complex features from artworks to build a better image representation. DCNN have outperformed handcrafted engineered features and has established the leading for the AMC. Nevertheless, BP has started to demonstrate its competence against DCNN performance in this area. However, AAs on art media represent a severe threat that has not been studied to measure the classifier's reliability.
Furthermore, the AA effect has not been demonstrated to influence different classification architectures' predictions, such as CV methods, with only one successful GP-like methodology \cite{Olague2020}. Although DCNN have developed defense mechanisms to diminished the AA effect, it is difficult to fight against all the new and more complex AA. Thus, even if DL architectures have classified large-scale sets of images with multiple classes with outstanding results, this paradigm's security concerns make the solutions unreliable. The brittleness is because, with small perturbations produced on the image, DL can be intentionally fooled. For example, there are critical areas in museums and galleries, such as artist identification and forgery detection, where the prediction's confidence must not depend on a system that can be manipulated by an imperceptible perturbation. This catastrophic scenario could lead to forgeries to circulate on the market or be misattributed to a specific artist. This article presents a method that can be used as a first defense mechanism by asking general questions like if the digitized artwork belongs to a certain class before asking further questions.

\section{Methodology}
\label{sec:Methodology}

This section describes the data modeling of each method used in this work. The main goal in data modeling is to summarize the data by fitting it to a model by establishing a relationship within the data $(y,\mathbf{x})$ given by the dataset by the following equation:
\begin{equation}
\label{eq:model}
    y=f(\mathbf{x}) \;\;\;,
\end{equation}
\noindent
where the function $f()$ is the model that depends on adjustable parameters. Therefore, we detail SIFT + Fisher Vectors modeling as the BoV method because it was the last computer vision technique that won the image classification task on the ImageNet Large Scale Visual Recognition Challenge (ILSVRC) 2011 before DL models arose. We describe the modeling from deep neural networks and explain the contributions to the state-of-the-art from the four DCNN models chosen for this work based on the ILSVRC winners. Next, we present the theory behind BP to introduce function-symbolic learning for data modeling and the workflow from the system. Finally, we describe the modeling from the three selected AAs to construct the perturbation to induce a misclassification such as in Equation (\ref{eq:aa}).

\subsection{SIFT + Fisher Vectors}

Fisher Vector (FV) is a vectorial representation of the gradient of the sample log-likelihood concerning a generative model of the data \cite{sanchez2013image}. There are many advantages to the FV against the BoV. It was proved by \cite{sanchez2013image} that BoV is a particular case of the FV where is restricted the gradient computation to the mixture weight parameters of the Gaussian Mixture Model (GMM) \cite{titterington1985statistical}. The generative model (GMM) can be understood as a probabilistic visual vocabulary. Nevertheless, FV incorporates additional gradients that improve accuracy. Also, it needs fewer vocabularies with lower computational costs than BoV, and it is easy to achieve good performance with simple linear classifiers. Note that BoV is typically quite sparse while the FV is almost dense, making FV impractical for large-scale applications due to storage problems. Nonetheless, a large-scale nearest neighbor search is made to mitigate this problem using a popular computer vision method named product quantization \cite{gray1998quantization}. In practice, it is used SIFT descriptors on a dense multi-scale grid to compute the FV image representation \cite{sanchez2013image}.

In order to construct the FV image representation, it is defined a set of D-dimensional descriptors extracted from an image $X = \{x_{t},t=1,\dots,T\}$, a set of SIFT descriptors. FV is a sum of normalized gradient statistics $\delta_{\lambda}^{X}=\sum_{t=1}^{T}L_{\lambda}\nabla_{\lambda}\log u_{\lambda}(x_{t})$ with the assumption that all descriptors are independent. Where $L_{\lambda}\nabla_{\lambda}\log u_{\lambda}(x_{t})$ is the normalized gradient statistics computed for each descriptor. Therefore, it can be understood that this operation is an embedding of the local descriptors $x_{t}\rightarrow \phi_{FK}(x_{t})=L_{\lambda}\nabla_{\lambda}\log u_{\lambda}(x_{t})$ in a higher-dimensional space which helps a linear classifier to model the data easier as in Equation (\ref{eq:model}).

These algorithms' advantage is that it does not require labeled data to learn the \textit{dictionary}. Therefore, it can work on limited labeled data situations. The dictionary learning process can also improve features quality by providing additional information of them \cite{zhang2010discriminative,jiang2011learning}. However, they are not capable of building features hierarchies, and the process is not merely stacked one method on the top of other even there have been attempts to make it deep \cite{coates2011analysis,simonyan2013deep,he2014unsupervised}.

\subsection{Deep Convolutional Neural Networks}

Differently, ANN starts the idea of designing deep architectures for neural network models that can extract sufficient features along this structure to allow the ANN to classify images. Deep Neural Networks, where DCNN is part of it, models the data using Equation (\ref{eq:model}) employing $f_{DNN}()$ as a particular form of a nested function, and each one called a layer.
\begin{equation}
    y=f_{DNN}(x)=f_{3}(\mathbf{f}_{2}(\mathbf{f}_{1}(\mathbf{x}))) \;\;\;,
\end{equation}
\noindent
in such a way that $\mathbf{f}_{1}$ and $\mathbf{f}_{2}$ are vector functions of the following form:
\begin{equation}
    \mathbf{f}_{\textit{l}}(\mathbf{z})=\mathbf{g}_{\textit{l}}(\mathbf{W}_{\textit{l}}\mathbf{z}+\mathbf{b}_{\textit{l}}) \;\;\;,
\end{equation}

\noindent
with \textit{l} denoting the index of the layer. $\mathbf{g}_{\textit{l}}$ is the activation function that usually is a nonlinear function, and the model parameters consist of $\mathbf{W}_{\textit{l}}$ the weights matrix and $\mathbf{b}_{\textit{l}}$ the bias vector. Hence, the minimization problem is defined by the loss function $J(\theta,\mathbf{x},y)$ where the goal is to find the best model parameters for all the layers $\Theta$ that fits the data $\mathbf{x}$ to the label $y$.

LeCun et al. \cite{lecun1989backpropagation} introduced the modern framework of Convolutional Neural Networks (CNN's). However, the first time that CNN starts attracting attention was with the development of the AlexNet \cite{krizhevsky2012imagenet}, a DCNN model for the ILSVRC 2012, where it could reduce by half the error rate on the image classification task. AlexNet layer architecture consists of 5 convolutional, three max-pooling, two normalizations, three fully connected layers (the last with 1000 softmax output), 60 Million parameters, and 500,000 neurons. Additionally, \cite{krizhevsky2012imagenet} introduced the use of ReLU (Rectified Linear Unit) Non-linearity as activation function with the benefits of much faster training than using \emph{tanh} or \emph{sigmoid} functions. To prevent overfitting, they also introduced the dropout method and data augmentation.

Another DL model that brought contributions to the state-of-the-art was the VGG network from the Visual Geometry Group of the University of Oxford \cite{simonyan2014very}. VGG network increased the deep of previous networks by creating VGG-16 and VGG-19. The first one uses 13 convolutional layers and three fully connected layers; the second one employed three additional convolutional layers. Also, they reduced the size of the filters to the smallest size to capture the notion of up/down, left/right, and center that is a 3x3 filter. VGG was distinguished for its state-of-the-art performance on recognition and localization tasks at ILSVRC 2014 and other image recognition datasets.

ResNet \cite{he2016deep} (Deep Residual Learning for Image Recognition) also contributed to redefining the layer as a residual learning function on the CNN architecture. This helps to mitigate the bottleneck problem of the training phase on CNNs. ResNet showed its capacity to train its architecture with a depth of up to 152 layers and lower complexity than GoogLeNet. Also, ResNet won the ILSVRC 2015 on the classification task achieving for the first time an error rate of 3.57\%. They proposed five configurations of the network: 18-layer, 34-layer, 50-layer, 101-layer, and 152-layer networks.

\subsection{Brain Programming}

Before we explain the algorithm of BP, we make a brief introduction to GP algorithms. GP is an evolutionary computation technique inspired by biological evolution principles \cite{Poli2008AFG}. It is considered a derivative of genetic algorithms that evolve individuals' populations in the form of a tree or computer program (formulas or mathematical expressions). Each individual computer program is generated depending on the terminal and function sets established by the user. They are evaluated in terms of how well it performs in a particular problem. Then, using the Darwinian principle of reproduction and survival of the fittest and the genetic operators of crossover and mutation, individuals are evolved to find a better fit solution to the problem.

BP is an evolutionary paradigm for solving CV problems reported in \cite{HERNANDEZ2016216,Olague2019,Olague2019b}. This methodology extracts characteristics from images through a hierarchical structure inspired by the brain's functioning. BP proposes a GP-like method, using a multi-tree representation for individuals. The main goal is to obtain a set of evolutionary visual operators ($EVOs$), also called visual operators ($VOs$), which are embedded within a hierarchical structure called the artificial visual cortex. The AVC is based primarily on two models: a psychological model called feature integration theory \cite{TREISMAN} and a neurophysiological model called the two pathway cortical model \cite{GOODALE1992}. Thus, the AVC attempts to emulate the natural process that occurs along the visual cortex according to the brain's neurological ventral-dorsal model. This two-streams model states that the process of acquiring visual information in the brain follows two main pathways. 

The dorsal stream is known as the ``where" or ``how" stream. This pathway is where the guidance of actions and recognizing objects' location in space is involved and where visual attention occurs. The first theory states that visual attention in human beings is performed in two stages. The first is called the preattentive stage, where visual information is processed in parallel over different feature dimensions that compose the scene: shape, color, orientation, spatial frequency, brightness, and motion direction. The second stage, called focal attention, integrates the extracted features from the previous stage to highlight a region of the scene. BP is based on the most popular theory of feature integration for the dorsal stream from \cite{TREISMAN}, and the principles of the first computational model for visual attention, where the image is decomposed into several dimensions to obtain a set of conspicuity maps, which are then integrated into a single map called the saliency map \cite{Koch1985ShiftsIS}.

The ventral stream is known as the ``what" stream. This pathway is mostly associated with object recognition and shape representation tasks. Proposed ventral stream models like neocognitron system \cite{fukushima1980biological}, convolutional neural networks \cite{lecun1989backpropagation}, and HMAX model \cite{Riesenhuber1999} (the Max principle is used in BP), start by decomposing the image into a set of alternating ``S" and ``C" layers. The ``S" or simple layers are defined by a set of local filters applied to find higher-order features, and the ``C" complex layers increase the features invariance by combining units of the same kind. However, BP replaces the data-driven models with a function-driven paradigm. In the function-driven process, a set of visual operators is fused by synthesis to describe the image's properties. Through a set of experiments, we will show that the discovered solutions do not rely directly on the data but specific characteristics; hence, making the solutions reliable regarding AAs.

Therefore BP can be summarized in two steps: first, the evolutionary process whose primary purpose is to discover functions to optimize complex models by adjusting the operations within them. Second, the AVC, a hierarchical structure inspired by the human visual cortex, uses the concept of function composition to extract features from images. The model can be adapted depending on the task, whether it is trying to solve the focus of attention for saliency problems or the complete AVC for categorization/classification problems. BP differs from the data-driven models using a function-driven approach to extract and combine the relevant information that solves a specific visual task. Hence, the overall function-driven process requires the input in a suitable representation; thus, we define an image $ I $ as the graph-of-a-function. 

\textbf{Definition 1. Image as the graph of a function}. \textit{Let $f$ be a function $f:U \subset \mathbb{R}^2 \rightarrow \mathbb{R}$. The graph or image $I$ of $f$ is the subset of $\mathbb{R}^3$ that consist of the points $(x, y, f(x,y))$, in which the ordered pair $(x,y)$ is a point in $U$ and $f(x,y)$ is the value at that point. Symbolically, the image $I = \{(x,y,f(x,y)) \in \mathbb{R}^3 | (x,y) \in U\}$}.

This definition is based on the fact that the images result from the impression of variations in light intensity along the two-dimensional plane. Therefore, functions are optimized to imitate the functionality of specialized areas of the brain through a set of operators.

\subsubsection{Data Modeling with BP}

BP proposes to solve the problem of image classification from the standpoint of data modeling through GP. Therefore, to understand the learning process of BP, we start defining the minimization problem, which requires to find a solution $\mathbf{P}_{min} \in S$ such that:

\begin{equation}
   \forall \mathbf{P}_{min} \in S : f(\mathbf{P}_{min}) \leq f(\mathbf{P}) \;\;\;.
\end{equation}

Hence, opposed to conventional approaches to finding the best-fit parameters, we would like to fit the data through discovering functions that perform a classification task in BP. The strategy takes several steps because the direct mapping between the domain and codomain is unknown or not well defined. In this manner, the solution to the image classification problem through BP requires to define the following equation:

\begin{equation}
    y= min(f(\mathbf{x},\mathbf{F},\mathbf{T},\mathbf{a})) \;\;\;,
    \label{eq:main}
\end{equation}
\noindent
where $(y,\mathbf{x})$ are the label and the image respectively, given by the dataset; $\mathbf{F}$ represents the set of functions, $\mathbf{T}$ defines the terminal set, and $\mathbf{a}$ are the parameters controlling the evolutionary process. Therefore, in order to solve the problem, we need two things: 1) a method of feature extraction and 2) a suitable criterion $\mathbf{Q}$ for the minimization.

Therefore, BP is the algorithm in charge of tuning $(\mathbf{F},\mathbf{T},\mathbf{a})$ looking for optimal feature extraction from the input images using the visual operators embedded into the artificial visual cortex (AVC). The criterion for the minimization $\mathbf{Q}$ in terms of a classification task helps discover the best classifier. In this particular case, we use an SVM to learn a mapping $f(\mathbf{x})$ that associates descriptors $\mathbf{x_i}$ to labels $y_i$. Here, we define the BP algorithm in terms of a binary classification task, whose main purpose is to find a decision boundary that best separates the class elements.

\subsubsection{Evolving an Artificial Visual Cortex (AVC)}

\begin{figure}
\includegraphics[width=\textwidth]{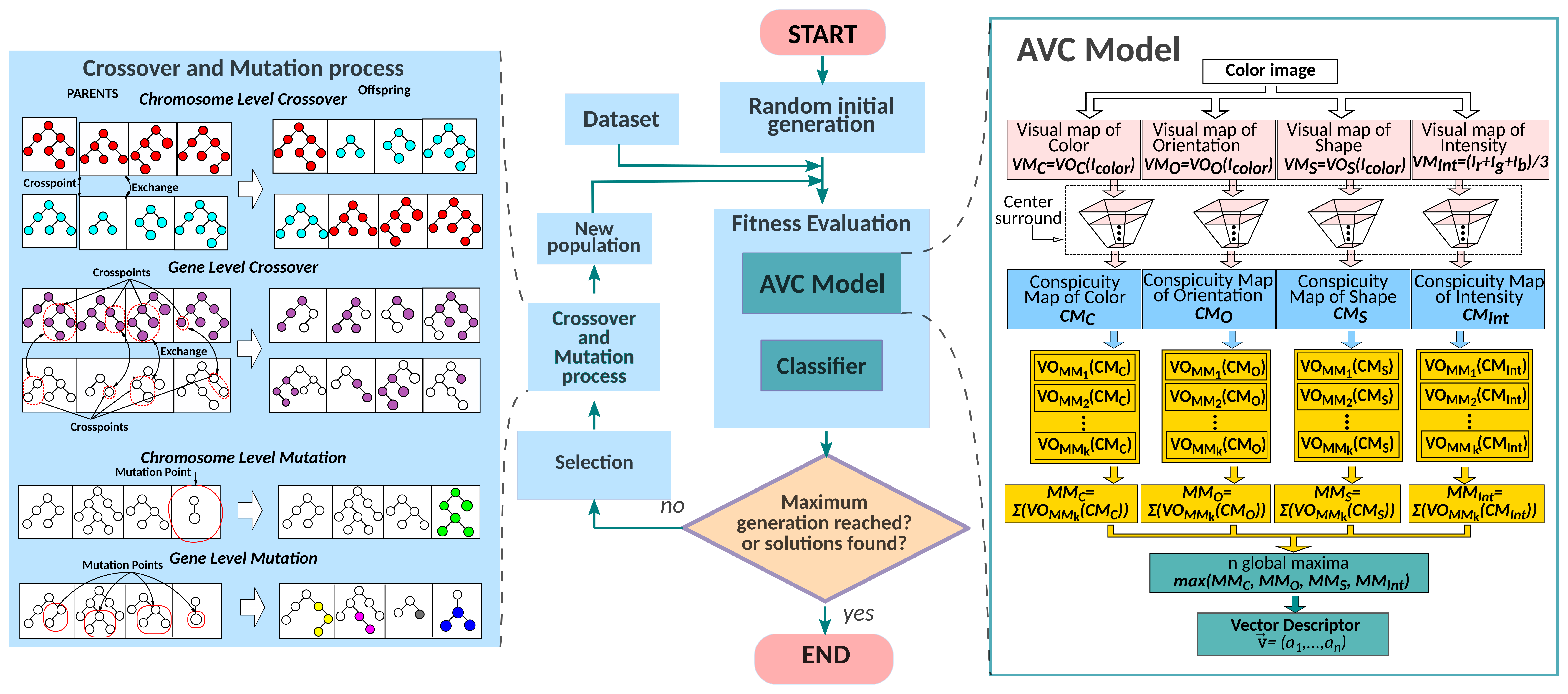}
\caption{Brain Programming workflow. The left side shows the genetic operations, in the middle is presented the BP's flow diagram, and the right side illustrates the individual representation.}
\label{fig:BP}
\end{figure}

Each individual consists of syntactic trees defining the $VOs$ that constructs the AVC structure to extract features from color images. This procedure gets a descriptor vector that encodes salient characteristics from the image. Then, we apply an SVM to calculate the classification accuracy for a given training image database to obtain the individual fitness. Hence, BP uses an evolutionary loop presented in Algorithm \ref{al:evolve} to evolve the entire population represented by a set of AVCs, in which the whole workflow is illustrated in Figure \ref{fig:BP}.

\IncMargin{1em}
\begin{algorithm}
\label{al:evolve}
\SetKwData{Left}{left}\SetKwData{This}{this}\SetKwData{Up}{up}
\SetKwFunction{Union}{Union}\SetKwFunction{FindCompress}{FindCompress}
\SetKwInOut{Input}{Input}\SetKwInOut{Output}{Output}
\Input{Training images, Algorithm parameters (see Table \ref{tab:parameters})}
\Output{The updated population AVCs}
\emph{Generate a random initial population $P_{0}$}\;
$i=0$\;
\While{the termination criterion is not satisfied}{
    Evaluate each individual fitness (AVC) in $P_{i}$ \;
    Selection using lexicographic parsimony pressure\;
    Generate offspring by crossover and mutation\;
    $i=i+1$\;
 }
 \textbf{return} The updated population $P_{final}$ 
\caption{BP evolutionary process}\label{algo_disjdecomp}
\end{algorithm}
\DecMargin{1em}

\subsubsection{Structure Representation and Genetic Operations}

In BP, an individual is a computer program represented by syntactic trees embedded into a hierarchical structure. Individuals within the population contain a variable number of syntactic trees, ranging from 4 to 12, one for each evolutionary visual operator ($VO_O$, $VO_C$, $VO_S$, $VO_I$) regarding orientation, color, shape, and intensity; and at least one tree to merge the resulting visual maps, and finally generate the Mental Maps (MM). All functions within each $VO$ are defined according to expert knowledge to highlight characteristics related to the respective feature dimension and updated through genetic operations.

\begin{itemize}
    \item Visual Maps 
\end{itemize}

Each input image is transformed to build the set $I_{color}$ = $\{I_r$, $I_g$, $I_b$, $I_c$, $I_m$, $I_y$, $I_k$, $I_h$, $I_s$, $I_v\}$, where each element corresponds to the color components of the RGB (red, green, blue), CMYK (Cyan, Magenta, Yellow, and black) and HSV (Hue, Saturation, and Value) color spaces. Elements on $I_{color}$ are the inputs to four $VOs$ defined by each individual. Each $VO$ is a mapping function applied to the input image to extract specific features from it, along with information streams of color, orientation, shape, and intensity; each of these properties is called a dimension. The output to $VO$ is an image called Visual Map ($VM$) for each dimension. It is important to note that each solution in the population should be understood as a complete system and not only as a list of three-based programs. Individuals represent a possible configuration for feature extraction that describes input images and is optimized through the evolutionary process. Next, we explain the process of $VOs$ to extract features on each dimension to obtain a resulting $VM$.

The first tree of the individual mimics the orientation. Thus, we evolve this visual operator ($ VO_O $) through a set of specially selected elements to highlight edges, corners, and other orientation-related features using the set of terminals and functions provided in Table \ref{tab:Ofunc}. The input for the functions can be any of the terminals, as well as the composition among them; $G_{\sigma}$ are Gaussian smoothing filters with $\sigma = 1,2$; and $D_u$ represents the image derivatives along the direction $u \in {x, y, xx, yy xy}$. These operators emulate the functionality of the V1 region presented in the primary visual cortex. 

\begin{table}[h!]
\caption{Functions and terminal list for the $VO_O$. }
\centering
\resizebox{\textwidth}{!}{
    \begin{tabular} {p{0.13\linewidth}p{0.25\linewidth}p{0.30\linewidth}p{0.15\linewidth}p{0.17\linewidth}}
      \toprule
       Dimension &  Functions  & Description & Terminals  & Description \\
      \midrule
      $VO_O$&$A+B$, $A-B$, $A \times B$, $A / B$, $k+A$, $k-A$, 
      $k \times A$, $A / k$, 
      $|A|$, $|A+B|$, $|A-B|$, $log(A)$, $(A)^2$, $\sqrt{A}$, 
      $round(A)$, $\lfloor A \rfloor $, $\lceil A \rceil $,
      $inf(A,B)$, $sup(A,B)$, $G_{\sigma=1}(A)$, $G_{\sigma=2}(A)$, $D_{x}(A)$, $D_{y}(A)$, $thr(A)$ &
      Arithmetic functions between images or constants $k$, absolute values, trascendental functions, square, square root, rounding functions, infimum, supremum, convolution with a Gaussian filter, derivatives, and threshold applied to images $A$ and/or $B$ &
      $I_r$, $I_g$, $I_b$, $I_c$, $I_m$, $I_y$, $I_k$, $I_h$, $I_s$, $I_v$, 
      $D_{x}(I_x)$, $D_{xx}(I_x)$, $D_{y}(I_x)$, $D_{yy}(I_x)$, $D_{xy}(I_x)$ & Elements of $I_{color}$ and its derivatives\\
      \bottomrule
     \end{tabular}
    }
\label{tab:Ofunc}
\end{table}

The second operator encodes the color dimension emulating the color-sensitive cells in the visual cortex. The visual operator of color($VO_C$) reproduces the color perception process to find prominent regions with color properties in the image. Note that some functions of $ VO_C $ are the same as those in $ VO_O $ plus the function $complement( )$ that provides a negative image that complements an intensity or RGB value (see Table\ref{tab:Cfunc}). Thus, in the output image, dark areas become lighter, and light areas become dark. Opponent terminals perform a fixed operation between the color bands that builds a new image with the maximum values between them. For example, $ Op _ {\ r, g}$ accentuates the difference between the red and green bands.

\begin{table}[h!]
\caption{Functions and terminal list for the $VO_C$. }
\centering
\resizebox{\textwidth}{!}{
    \begin{tabular} {p{0.13\linewidth}p{0.25\linewidth}p{0.30\linewidth}p{0.15\linewidth}p{0.17\linewidth}}
      \toprule
       Dimension &  Functions  & Description & Terminals  & Description \\
      \midrule
      $VO_C$ & $A+B$, $A-B$, $A \times B$, $A / B$, $k+A$, $k-A$, 
      $k \times A$, $A / k$, $log(A)$, $exp(A)$, 
      $(A)^2$, $\sqrt{A}$, $(A)^c$, $round(A)$,
      $\lfloor A \rfloor $, $\lceil A \rceil $, $thr(A)$ &  
      Arithmetic functions between images or constants $k$, trascendental functions, square, square root, image complement, rounding functions and threshold applied to images $A$ and/or $B$ & $I_r$, $I_g$, $I_b$, $I_c$, $I_m$, $I_y$, $I_k$, $I_h$, $I_s$, $I_v$, $Op_{r-g}(I)$, $Op_{b-y}(I)$  & Elements of $I_{color}$ and color opponencies: red-green and blue-yellow\\ 
      \bottomrule
     \end{tabular}
    }
\label{tab:Cfunc}
\end{table}

The third tree is the visual operator of shape. The method that extracts visual information from the object's shape employing $ VO_S $ from Table\ref{tab:Sfunc}, which utilize the morphological information of the artifacts in the image. BP proposes to create compound operators by the composition of basic morphological operators such as dilation, erosion, open, close with disk, square, and diamond structural elements. Indeed, more complex operators can be created from these operators. The goal of extracting shape information is to highlight morphological information that can be used for object recognition.

\begin{table}[h!]
\caption{Functions and terminal list for the $VO_S$. }
\centering
\resizebox{\textwidth}{!}{
    \begin{tabular} {p{0.13\linewidth}p{0.25\linewidth}p{0.30\linewidth}p{0.15\linewidth}p{0.17\linewidth}}
      \toprule
       Dimension &  Functions  & Description & Terminals  & Description \\
      \midrule
      $VO_S$ & $A+B$, $A-B$, $A \times B$, $A / B$, $k+A$, $k-A$, $k \times A$, $A / k$, 
      $round(A)$, $\lfloor A \rfloor $, $\lceil A \rceil $, $thr(A)$, $A \oplus SE_d$, 
      $A \oplus SE_s$, $A \oplus SE_{dm}$, $A\ominus SE_{d}$, $A\ominus SE_{s}$, 
      $A\ominus SE_{dm}$, $A\circledcirc SE_s$, $A\odot SE_s$, $Sk(A)$, $Perim(A)$, $A\circledast SE_{d}$, $A\circledast SE_{s}$, $A\circledast SE_{dm}$, $T_{hat}(A)$, $B_{hat}(A)$  & 
      Arithmetic functions between images or constants $k$, rounding functions, threshold, and morphological operators: dilation, erosion, open, close with disk, square, and diamond structural element; skeleton, hit or miss, bottom-hat, top-hat & 
      $I_r$, $I_g$, $I_b$, $I_c$, $I_m$, $I_y$, $I_k$, $I_h$, $I_s$, $I_v$ & Elements of $I_{color}$\\ 
      \bottomrule
     \end{tabular}
    }
\label{tab:Sfunc}
\end{table}

Finally, the intensity measure corresponds to the amount of light perceived by a photosensitive device.  In humans, the intensity is measured by specialized ganglion cells in the retina. Then, the following formula is applied to compute the visual map of intensity.
\begin{equation}
VM_{Int} = \frac{I_r + I_g + I_b}{ 3 } \;\;\; .   
\end{equation}

\begin{itemize}
    \item Conspicuity Maps
\end{itemize}
 
The next procedure is the center-surround process; it efficiently combines the information from the $VMs$ and is useful for detecting scale invariance in each of the dimensions. This process is performed by applying a Gaussian smoothing over its corresponding $VM_d$ at nine scales $P^\sigma_d = \{ P^{\sigma=0}_d, P^{\sigma=1}_d, ..., P^{\sigma=7}_d, P^{\sigma=8}_d\}$; this processing reduces the visual map's size by half on each level forming a pyramid. Subsequently, the six levels of the pyramid are extracted and combined. 

\begin{equation}
Q^j_d = P_d^{\sigma=\lfloor \frac{j+9}{2} \rfloor+1} - P_d^{\sigma=\lfloor \frac{j+2}{2} \rfloor+1} \;\;\;,
\end{equation}

\noindent
where $j = {1, 2, . . . , 6}$. Since the levels $P^\sigma_d$ have different sizes, each level is normalized and scaled to the visual map's dimension using polynomial interpolation. This technique simulates the center-surround process of the biological system. After extracting features, the brain receives stimuli from the vision center and compares it with the receptive field's surrounding information. The goal is to process the images so that the results are independent of scale changes. The entire process ensures that the image regions are responding to the indicated area. This process is carried out for each characteristic dimension ($VM_d$); the results are called Conspicuity Maps ($CM$), focusing only on the searched object by highlighting the most salient features. This early stage of the system follows the psychological model of visual attention, which involves the objects' location in space as the artificial dorsal stream pathway.

\begin{itemize}
    \item Mental Maps
\end{itemize}

After obtaining the most saliency features, the next stage along the AVC is to compute the Mental Maps ($MMs$) to define a descriptor vector used as input to a classifier for categorization purposes. This procedure is analogous to the artificial ventral stream pathway. Hence, the information from $CMs$ is synthesized to build the set of $MMs$, which discriminates unwanted information. The AVC model uses a set-of-functions to extract the images' discriminant characteristics (see Table \ref{tab:MMfunc}); it uses a functional approach. Thus, a set of $k$ $VOs$ is applied to the $CMs$ for the construction of the $MMs$. These $VOs$ correspond to the remaining part of the individual that has not been used. Unlike the operators used for the $VMs$, the operators' whole set is the same for all the dimensions. These operators filter the visual information and extract the information that characterizes the object of interest. Then, using Equation (\ref{eq:MM}), where $d$ is the dimension, and $k$ represents the cardinality of the set of $VO_{MM_k}$, we apply the $MMs$ for each dimension.

\begin{equation} \label{eq:MM}
MM_d = \sum_{i=1}^k VO_{MM_i}\left(CM_d\right)
\end{equation}

\begin{table}[h!]
\caption{Functions and terminal list for the set $VO_{MM}$. }
\centering
\resizebox{\textwidth}{!}{
    \begin{tabular} {p{0.13\linewidth}p{0.25\linewidth}p{0.30\linewidth}p{0.15\linewidth}p{0.17\linewidth}}
      \toprule
       Dimension &  Functions  & Description & Terminals  & Description \\
      \midrule
      $VO_{MM}$ & $A+B$, $A-B$, $A \times B$, $A / B$, $|A+B|$, $|A-B|$, $log(A)$, $(A)^2$, 
      $\sqrt{A}$, $G_{\sigma=1}(A)$, $G_{\sigma=2}(A)$, $D_{x}(A)$, $D_{y}(A)$  &
      Arithmetic functions between images or constants $k$, absolute values, transcendental functions, square, square root, convolution with a Gaussian filter, and derivatives & $CM_d$, $D_{x}(CM_d)$, $D_{xx}(CM_d)$, $D_{y}(CM_d)$, $D_{yy}(CM_d)$, $D_{xy}(CM_d)$ & Conspicuity Maps and its derivatives \\
      \bottomrule
     \end{tabular}
    }
\label{tab:MMfunc}
\end{table}

\begin{itemize}
    \item Genetic Operations
\end{itemize}

Individuals are selected from the population using a proportionate fitness method called lexicographic parsimony pressure to participate in the genetic recombination from the individuals' multi-tree representation. This method consists of assigning to each solution a selection probability proportional to their fitness value while preferring smaller trees when fitness is equal. The best individuals are retained to apply genetic operators to create the new offspring.  

Like genetic algorithms, BP executes the crossover between two selected parents at the chromosome level using a ``cut-and-splice" crossover. Thus, all data beyond the selected crossover point is swapped between both parents A and B. The result of applying a crossover at the gene level is performed by randomly selecting two subtree crossover points between both parents. The selected genes are swapped with the corresponding subtree in the other parent. The chromosome level mutation leads to selecting a given parent's random gene to replace such substructure with a new randomly mutated gene. The mutation at the gene level is calculated by applying a subtree mutation to a probabilistically selected gene; the subtree after that point is removed and replaced with a new subtree. These genetic operators allow the variation of the genetic material while promoting individuals' genetic innovation through all levels and maintaining the diversity of the population. 

\subsubsection{Fitness Function}

The following stage in the model is the construction of the image descriptor vector ($DV$). The system concatenates the four $MMs$ and uses a max operation to extract the $n$ highest values; these values are used to construct the $DV$. Once we get the DVs from images in the database, a classifier associates the domain given by the descriptors to the labels' codomain. In this work, we use an SVM working with the discriminate hyperplane defined by:

\begin{equation}
       f\left(\bm{x}\right) = \sum_{i=1}^{l} \alpha_i y_i K\left(\bm{x}_i,\bm{x}\right) + b ,
\end{equation}
 
\noindent
where the given training data is $(\mathbf{x}_i,y_i)$, $i = 1,\ldots,l$, $y_i \in \{ -1,1\}$, $\mathbf{x}_i \in \mathbf{R}^p$ and $K(\mathbf{x}_i,\mathbf{x})$ is the kernel function. The sign of the output indicates the class membership of $\mathbf{x}$. Thus, finding the best hyperplane is performed through an optimization process that locate the margin between the class and non-class as the search criteria. Therefore, the minimization problem on the learning pentuple from Equation (\ref{eq:main}) remains as $\left(\left(\boldsymbol{x},\boldsymbol{y}\right), \boldsymbol{F}, \boldsymbol{T}, \boldsymbol{a}, \boldsymbol{Q} \right)$. Thus, the accuracy obtained by the SVM indicates the fitness of the individual \footnote{The accuracy denoted in this Section has the purpose of optimizing BP; nevertheless, the accuracy indicated in Section \ref{sec:Metrics} refers to the metric to measure the attack responses.}.

\subsubsection{Initialization, GP parameters, and Solution Designation}

Once we define the AVC structure from each individual, we set the parameters of the evolutionary process of BP (see Table \ref{tab:parameters}) and establish the image database. Next, a random initial population is created using a ramped half-and-half technique, which selects half of the individuals with the grow method and half with the full method. The full method makes balanced trees according to the maximum initial depth, while the grow method makes unbalanced trees allowing branches of varying lengths. Here we set a limit of maximum depth to avoid uncontrolled growth of trees over time. Tree depth is dynamically set using two maximum values to limit any individual's size within the population. The dynamic max depth is a maximum value that may not be surpassed unless the individual's fitness is better than the best solution found so far. If it occurs, the dynamic max depth value is updated to the new fittest individual. The real max depth is a hard limit that no individual may surpass under any circumstances. Selection is carried out using a tournament with lexicographic parsimony pressure while keeping the best individual. Finally, the evolutionary process is terminated until one of these two conditions is reached: 1) an acceptable classification rate or 2) the total number of generations. Thus, the evolutionary process reaches an optimum population that contains the best solution to the problem.

\begin{table}[h!]
\caption{Initialization parameters for each GP applied in the BP algorithm.}
\begin{center}
\scalebox{0.8}{
\begin{tabular}{ll}
    \toprule
    Parameters                       & Description             \\ 
    \midrule
    Generations                      & 30                      \\ 
    Initial Population               & 30                      \\ 
    Crossover at chromosome level    & 0.4                     \\ 
    Crossover at gene level          & 0.4                     \\ 
    Mutation at chromosome level     & 0.1                     \\ 
    Mutation at gene level           & 0.1                     \\ 
    Tree depth                       & Dynamic depth selection \\ 
    Dynamic max depth                & 7 levels                \\ 
    Real max depth                   & 9 levels                \\ 
    Selection                        & Tournament with lexicographic \\
                                     & parsimony pressure \\ 
    Survival                         & Elitism          \\ 
    \bottomrule
\end{tabular}
}
\end{center}
\label{tab:parameters}
\end{table}

\subsection{Adversarial Attacks}
Adversarial attacks are classified depending on the model's available information and the desired attack to predict a specific class. Hence, we choose three different attacks: a white box untargeted (FGSM), a black box untargeted (one pixel attack), and a targeted attack (Adversarial Patch), which will be explained in the following paragraphs.

\subsubsection{Fast Gradient Sign Method}
\begin{figure}
\centering
\includegraphics[scale=0.7]{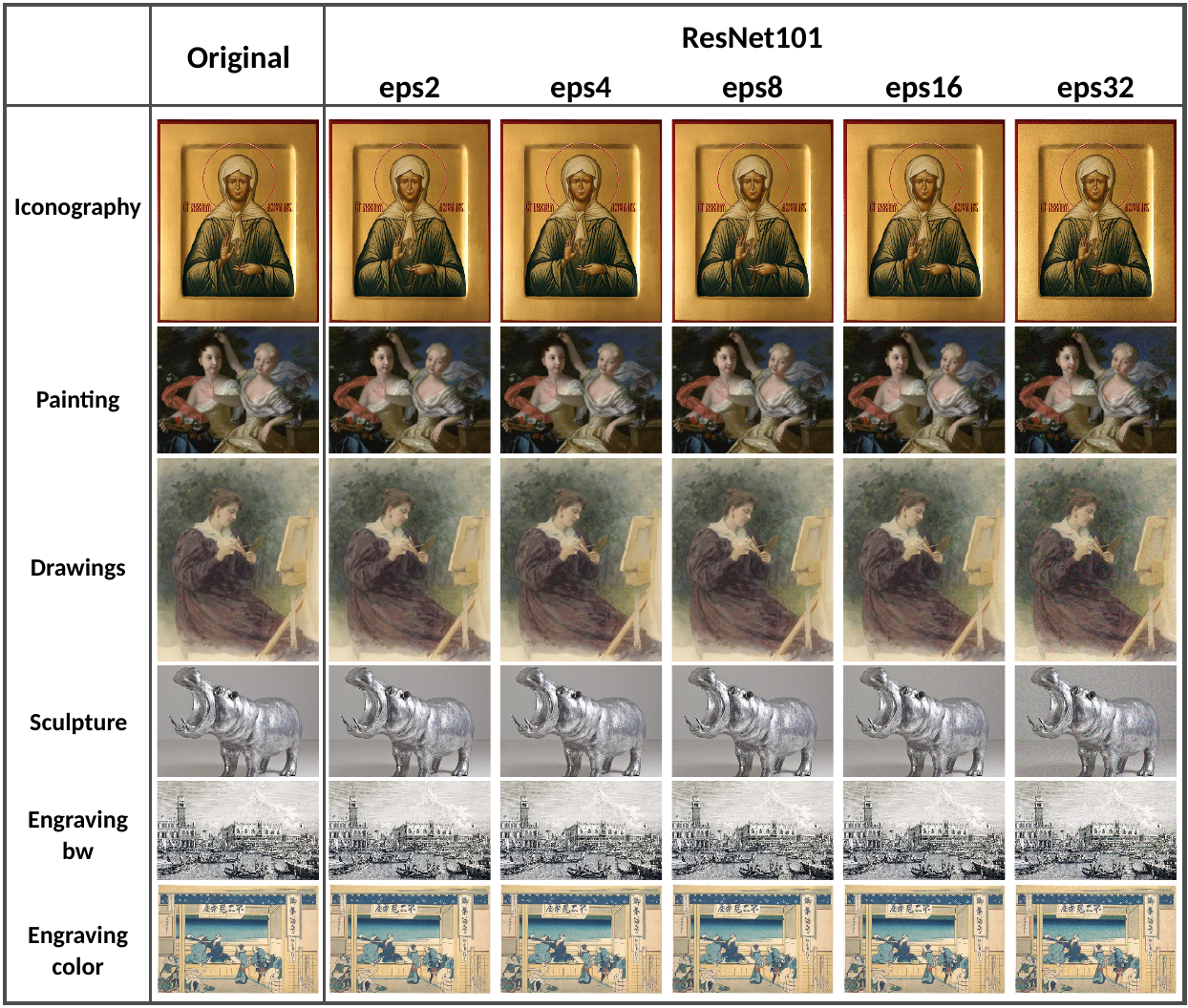}
\caption{Example images of computing the FGSM using ResNet101 from each class with a scale factor of $\epsilon={2,4,8,16,32}$.}
\label{fig:fgsm}
\end{figure}

The Fast Gradient Sign Method proposed by \cite{Goodfellow2015ExplainingAH}, is the most widely used method for computing AEs given an input image due to its easy implementation (See example images in Figure~\ref{fig:fgsm}). It proposes to increase the loss of the classifier by solving the following equation: $\rho = \epsilon\; \sign(\nabla J(\theta,\mathbf{x},y_l))$, where $\nabla J()$ computes the gradient of the cost function around the current value of the model parameters $\theta$ with the respect to the image $\mathbf{x}$ and the target label $y_l$. $\sign()$ denotes the sign function, which ensures that the magnitude of the loss is maximized and $\epsilon$ is a small scalar value that restricts the norm $L_{\infty}$ of the perturbation. 

The perturbations generated by FGSM take advantage of the linearity of the DL models in the higher dimensional space to make the model misclassify the image. The implication of the linearity of DL models discovered by FSGM is that exists transferability between models. Authors in \cite{Kurakin2017AdversarialML} reported that with the ImageNet dataset, the top-1 error rate using the perturbations generated by FGSM is around 63-69\% for $\epsilon \in \left[2, 32\right]$.  

\subsubsection{One Pixel Attack}

The one pixel attack was planed in a minimal scenario where only one pixel is changed in the image to fool the DL models using images of a reduced size of $32 \times 32$ pixels. With these limitations, Su et al. successfully fool three different CNN models on 70.97\% of the testing images with the modification of just one pixel per image \cite{su2019one}. Also, it was reported that the average confidence of the CNNs on the wrong prediction on the pictures was 97.47\%.

The one pixel adversarial perturbations are based on a black-box attack, on which no information about the network is required. It uses a population-based optimization algorithm for solving complex multi-modal optimization problems named Differential Evolution \cite{das2010differential} to generate the attack. It searches a solution from a vector space $\R^{5}$ that contains (\textit{x,y}) coordinates limited by the image size and the three bands of the RGB color values. Within a population, it randomly modifies the five-dimensional individuals' elements to create new offspring such that they compete in the current iteration to obtain better fitness. In the case of two pixels, an individual has a vector space $\R^{10}$ that contains the coordinates and colors values of both pixels, and so on for individuals with more pixels. During the run, the algorithm used the probability of the predicted label to compute the fitness criterion. The last surviving individual is used to modify the pixels in the image.

In summary, let the vector $\mathbf{x}=(x_{1},\dots,x_{n})$ be a \textit{n}-dimensional image, which is the input of the target classifier $f$ that predict correctly the class \textit{t} from the image. The probability of $\mathbf{x}$ associated to the class \textit{t} is $f_{l}(\mathbf{x})$. It builds an additive adversarial perturbation vector $e(\mathbf{x})=(e_1,\dots,e_n)$ according to $\mathbf{x}$, the class \textit{target} and the limitation of maximum modifications \textit{d}, a small number that express the dimensions that are modified while other dimensions of $e(\mathbf{x})$ left as zeros. For targeted attacks, the main purpose is to find the optimal solution $e(\mathbf{x})*$ that solves the following equation:
\begin{equation}
\begin{aligned}
    \max_{e(\mathbf{x})*} \quad &  f_{target}(\mathbf{x}+e(\mathbf{x}))\\
    \textrm{s.t.} \quad &  ||e(\mathbf{x})||_{0}\leq d \;\;\;.
\end{aligned}
\end{equation}

\begin{figure}
\centering
\includegraphics[scale=0.6]{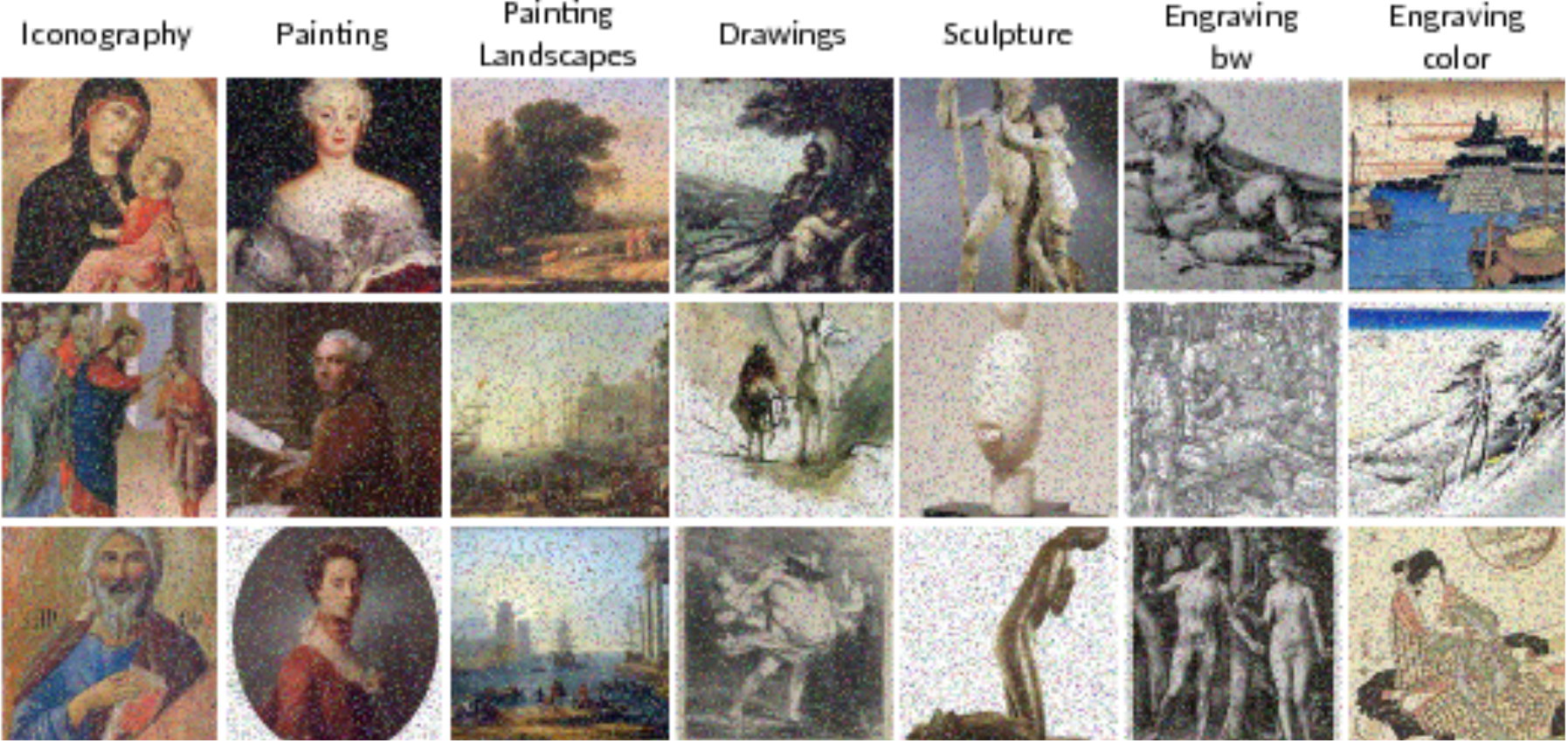}
\caption{Example images of the multiple pixel attack using $d=10,000$ for each class. Each column shows three sample images from the Wikiart database.}
\label{fig:onepixelattack}
\end{figure}

\noindent
Hence, the case of one pixel attack is $d=1$, but it can be extended to multiple pixels by increasing $d$. It should be noticed that one pixel attack was performed on DL models with inputs from CIFAR 10 dataset. So, it represents a considerable modification of such tiny images; nevertheless, it is insignificant with the databases studied in the present work. Therefore, we use a multiple pixel attack $d>>1$ in order to work with real size images. It should be noted that increasing the number of pixels in this attack will raise the perturbation risk to be noticeable (See example images in Figure~\ref{fig:onepixelattack}).

\subsubsection{Adversarial Patch}

\begin{figure}
\centering
\includegraphics[scale=0.55]{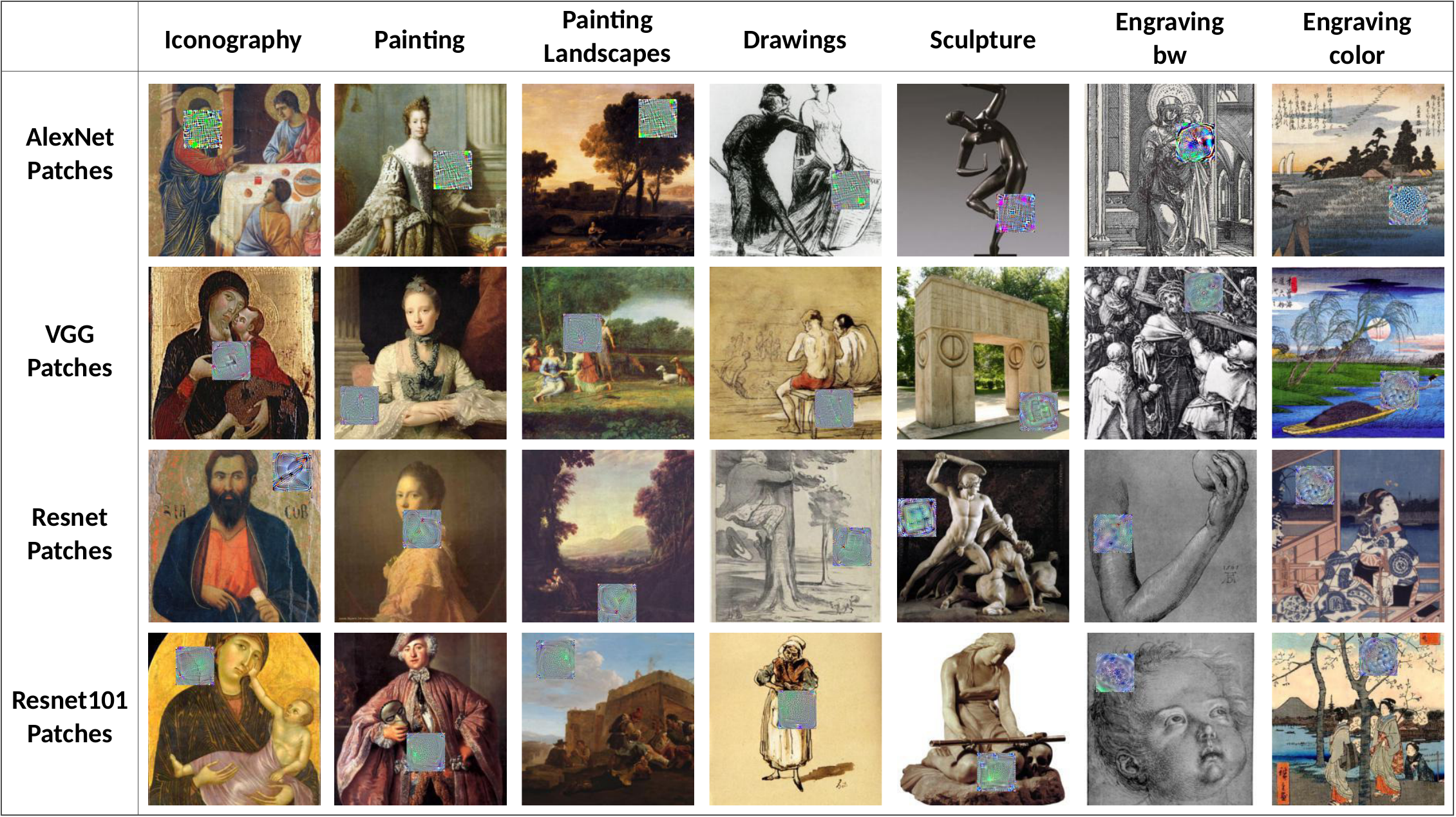}
\caption{Example images of the adversarial patch. Each column represents the classes from the Wikiart database, and each row represents the corresponding patch from the DCNN model.}
\label{fig:patches}
\end{figure}

The Adversarial Patch opposed to traditional strategy for creating a targeted AE by finding a maximum perturbation $e(\mathbf{x})$ that maximizes the $f_{target}(\mathbf{x}+e(\mathbf{x}))$ is a method to replace a perturbation on the whole image with a patch (See Figure~\ref{fig:patches}). The robustness of these patches resides on the wide variety of transformations on which they can attack any image and target the classifier prediction to the desired class. Also, they work in real work environments where they can be printed, photographed, or even when the patch is too small; they can make to ignore the whole scene to predict the target class.

To build patch $\hat{p}$, it was used a variant of the Expectation over Transformation (EOT) framework, on which the patch is trained to optimize the following equation:

\begin{equation}
    \hat{p}= \argmax_{\hat{p}} \E_{x\in X. t \in T. l \in L}[\log f(y,A(p,\mathbf{x},l,t))] \;\;\;,
\end{equation}

where $X$ is a training set of images, $T$ is a distribution over transformations of the patch, $L$ is a distribution over locations in the image, and $(y,\mathbf{x})$ are the label and the image vector respectively. The expectation over the training images improves the patch's effectiveness, regardless of what is in the background. It was proved by \cite{brown2017adversarial} the patch's universality using several images with different backgrounds. A variation of this method is to add a constraint of the form $||p - p_{orig}||_{\infty} < \epsilon$ to the patch objective in order to camouflage it. The constraint enforces the final patch to be within $\epsilon$ in the $L_\infty$ norm of some starting patch \textit{$p_{orig}$}.

\section{Experiments}
\label{sec:Experiments} 

Robust classification is a highly valuable characteristic regarding automatic system development following security and confidence of art pieces' predictions. In this study, we analyze the algorithms' performance using accuracy. Besides, we use the accuracy ratio between adversarial examples and clean images to measure robustness. Moreover, we propose a statistical analysis of each classifier's predictions' confidence to corroborate the results. Therefore, this experiment consists of studying the accuracy and robustness against AAs using three of the main approaches for image classifications: 
\begin{itemize}
    \item traditional handcrafted features algorithm (SIFT+FV)
    \item Deep Genetic Programming Methodology (BP)
    \item DCNN models (AlexNet, VGG, ResNet18, and ResNet101).
\end{itemize}

We consider unconventional training, validation, and test datasets since we apply two different image databases compiled by experts for AMC. Training and validation datasets are constructed from the Kaggle database, while testing uses a standard database WikiArt (See Table~\ref{Tab:dataset}). The aim is to emulate a real-world scenario where the proposed models are tested with standard benchmarks. 

This work analyzes the threat of using three types of AA to the model mentioned above. The white box untargeted (FGSM) determines the impact from an easy and direct threat to DCNN by knowing its parameters. Also, we study the transferability effect on other DCNN models, extending to BP and SIFT+FV, which are different architectures. We analyze the behavior of such perturbations from these architectures, which can cause wrong predictions with the addition of subtle texture to the artworks. The black box untargeted (multiple pixel attack) analyzes the hazard from an attack that tries to find locations and pixel values to build a perturbation that changes the model's prediction from an artwork image. The targeted attack uses the adversarial patch to challenge the robustness of such modified image patches, which can be rotated, put on random locations, and printed to appear in real-world conditions in the artwork to cause a misleading prediction of the target class. Additionally, we analyze the transferability effect of such patches through all models.

\subsection{Datasets}

We use the same datasets from the experiment of AMC reported in \cite{ChanLey2020CategorizationOD}. The training and validation set of images are obtained from the Kaggle website from the digitized artwork dataset. This dataset comprises five categories of art media: drawing, painting, iconography, engraving, and sculpture. The engraving class consists of two different kinds; most of them were black and white art pieces. The other style was Japanese engravings, which introduce color to the images. So, the engraving class was split into engraving black and white and engraving color. It is used for testing a standard database WikiArt, where it was selected the images from the same categories. Since the Wikiart engraving class is grayscale, the ukiyo-e class (Japanese engravings) from Wikiart was used as the engraving color class. Also, the set of images of the category landscapes, which are painting from renowned artists, is added to test the painting class. Table~\ref{Tab:dataset} provides the number of artworks for each dataset.

\begin{table}[H]
\caption{Total number of images per class obtained from Kaggle and Wikiart databases}
\begin{center}
\resizebox{\textwidth}{!}{
\begin{tabular}{p{0.15\linewidth}p{0.14\linewidth}p{0.14\linewidth}p{0.14\linewidth}p{0.14\linewidth}p{0.14\linewidth}p{0.14\linewidth}p{0.14\linewidth}}
\hline
 		& Iconography  & Painting & Drawings & Sculpture & Engraving BW  & Engraving Color& Caltech Background  \\ \hline  \hline
Train & 1038 & 1021 & 553 & 868 & 426 & 30 & 233\\ \hline  \hline
Validation & 1038 & 1021 & 553 & 868 & 283 & 19 & 233\\ \hline
Wikiart & 251 & 2089 & 204 & 116 & 695 & 1167 & 233\\ \hline 
Wikiart Landscapes &&136&&&&&\\ \hline

\end{tabular}
}
 \end{center}
\label{Tab:dataset}
\end{table}

\subsection{Evaluation metrics}
\label{sec:Metrics}
We employ classification accuracy as a measure of performance for the classifiers, which is simply the rate of correct classifications given by the following formula:
\begin{equation}
    Accuracy= \frac{1}{N}\sum_{n=1}^{N} d(y^{\prime}_{n},y_{n}) \;\;\;,
\end{equation}

\noindent
where $N$ is the total of test images, $y^{\prime}_{n}$ is the predicted label for the image $n$, $y_{n}$ is the original label for the image $n$, and $d(x,y)=1$ if $x=y$ and 0 otherwise.

Additionally, as a robustness measure, we used the accuracy ratio between adversarial examples and clean images implemented by \cite{Kurakin2017AdversarialML}. This metric means that if the ratio reaches one, the accuracy of AEs and the clean images is the same. Nevertheless, if it tends to zero, that means that the AA worked to fool the classifier. If this ratio exceeds 1, it implies that the AA is helping to correct misclassified images. The following equation calculates the ratio:
\begin{equation}
    Ratio=\frac{acc_{adv}}{acc_{clean}} \;\;\;,
\end{equation}
where $acc_{adv}$ is the classification accuracy on AEs, and $acc_{clean}$ is the classification accuracy on the clean images.

\subsection{Implementation details}
In this subsection, we outline the implementation details for all learned models:
\begin{itemize}
    \item Brain Programming: was implemented on Matlab using a modified version of GP Lab and the libsvm library for the SVM.
    \item SIFT+FV: was implemented on Matlab using VLFeat libraries for the SIFT description, GMM, and Fisher Vectors. It was used the SVM provided by Matlab.
    \item DCNN: for the implementation of the four models (AlexNet, VGG, ResNet18, and ResNet101), we use the pre-trained models from PyTorch v1.1. These models were retrained using transfer learning for the art media problem.
\end{itemize}

Also, we outline each of the AA:
    
\begin{itemize}  
    \item FGSM: was implemented in PyTorch v1.1 using the validation and test datasets to compute AEs with standard values for scale $\epsilon={2,4,8,16,32}$ for all the DCNN models.
    \item Multiple pixel attack: was implemented using 100 random images from the test dataset (50 from each class) in Matlab and Python. Python version was programmed using the differential evolution with the Pygmo library, and Matlab's version used the differential evolution library available from their file exchange website. Both implementations used the same settings of 50 individuals, 30 generations, a crossover probability of 0.9, and $d=10,000$ pixels.
     \item Adversarial Patch: was implemented using 100 images from the training dataset for each DCNN model in PyTorch v1.1 with the following parameters set to build the patch: patch size of $50 \times 50$ pixels, a max of 100 iterations per image with a stop criteria of 0.9 posterior probability of the target class. As we defined the binary classification problem, we choose the background class as the target prediction to measure the number of class images that predict the model as the target class.
\end{itemize}

\subsection{Results}
The results obtained from the experiments mentioned above are presented and discussed in the following subsection.

\subsubsection{FGSM}

In Table~\ref{tabkaggle}, we present the results for the training and validation datasets from Kaggle along with the AEs computed using FGSM for all DCNN models. We report classification accuracy at each stage of training and validation next to all models' accuracy tested with the AEs. Here, we want to measure the influence in the prediction of the FGSM in two manners: 1) direct, since we know the model's parameters and perturbation, and 2) indirect, through the transferability of the attack. Previously, other researchers reported that AEs could affect different CNN models by setting them up for the same task. Still, we want to extend the analysis to different architectures such as BP and SIFT+FV that could be affected by these subtle perturbations to the digitized artworks.

First, we observed that SIFT+FV models appeared to be overfitted. Hence, we perform two types of verifications presented in Table~\ref{crossvalidation}. We use the hyperparameters optimizer from Matlab and the \emph{crossval} function that validates the model using a 10-fold cross-validation. After ten runs, the hyperparameters optimizer returns the best model for each class. The results over the train and validation datasets are listed in the \emph{optimizer} column at Table~\ref{crossvalidation}. The \emph{crossval} function randomly partition the data into ten sets of equal size, later train an SVM classifier on nine sets, and repeat the process ten times. After that, we computed the mean accuracy at train and validation datasets for each class over each ten models. We present the results in the \emph{cross-validation} column at Table~\ref{crossvalidation}. We obtained the same results as the original experiment. Then, the results showed that the data do not overfit the models.

Therefore, it can be observed in Table~\ref{tabkaggle} how drastically can be dropped the performance of DCNN. The worst-case was the sculpture class, the VGG's performance went from 97.62\% to 14.38\%, AlexNet dropped from 95.78\% to 14.57\%, ResNet18 diminished from 96.88\% to 19.07\%, and ResNet101 decreased from 97.89\% to 37.86\%. Also, it is perceived that the transferability effect between the DCNN models is more significant at $\epsilon=32$. The drawings class presents almost the same behavior as the sculpture class, where the other networks are affected by AEs. For all other classes, the effect is unnoticeable, but the accuracy is significantly affected when the model matches the AE.

\begin{figure}
\centering
\includegraphics[width=0.7\textwidth]{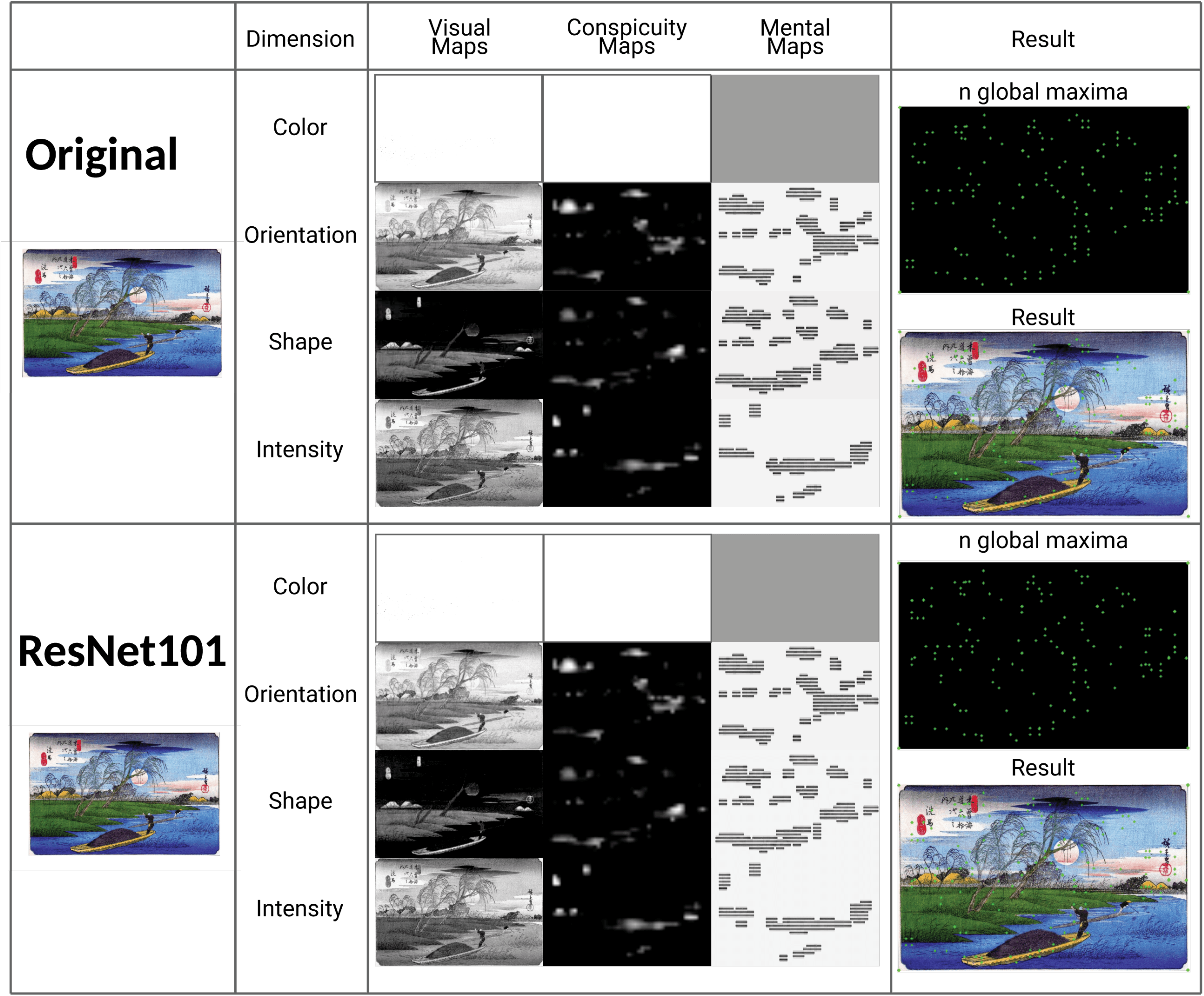}
\caption{Maps generated in each phase of the AVC, extracted from the original image and the AE computed from FGSM using ResNet101 with $\epsilon=32$. Note that despite the attack, the generated maps do not change much.}
\label{fig:dimensionResnet}
\end{figure}

In some cases, SIFT+FV was affected by FGSM. For example, in the drawing class, the performance was reduced by almost 8\%. And for the painting, the accuracy was decreased approximately by 4\%. This result shows a partial transferability of AEs to SIFT+FV because regardless of applying DCNN, the perturbation compromised these two classes' performance. However, BP maintains its performance in almost every test; the accuracy variation through all the analysis was less than 2\%. Figure \ref{fig:dimensionResnet} illustrates an example showing that the generated maps from the AVC do not suffer any change in their responses with the FGSM.  

\begin{figure}
\centering
\includegraphics[width=0.95\textwidth]{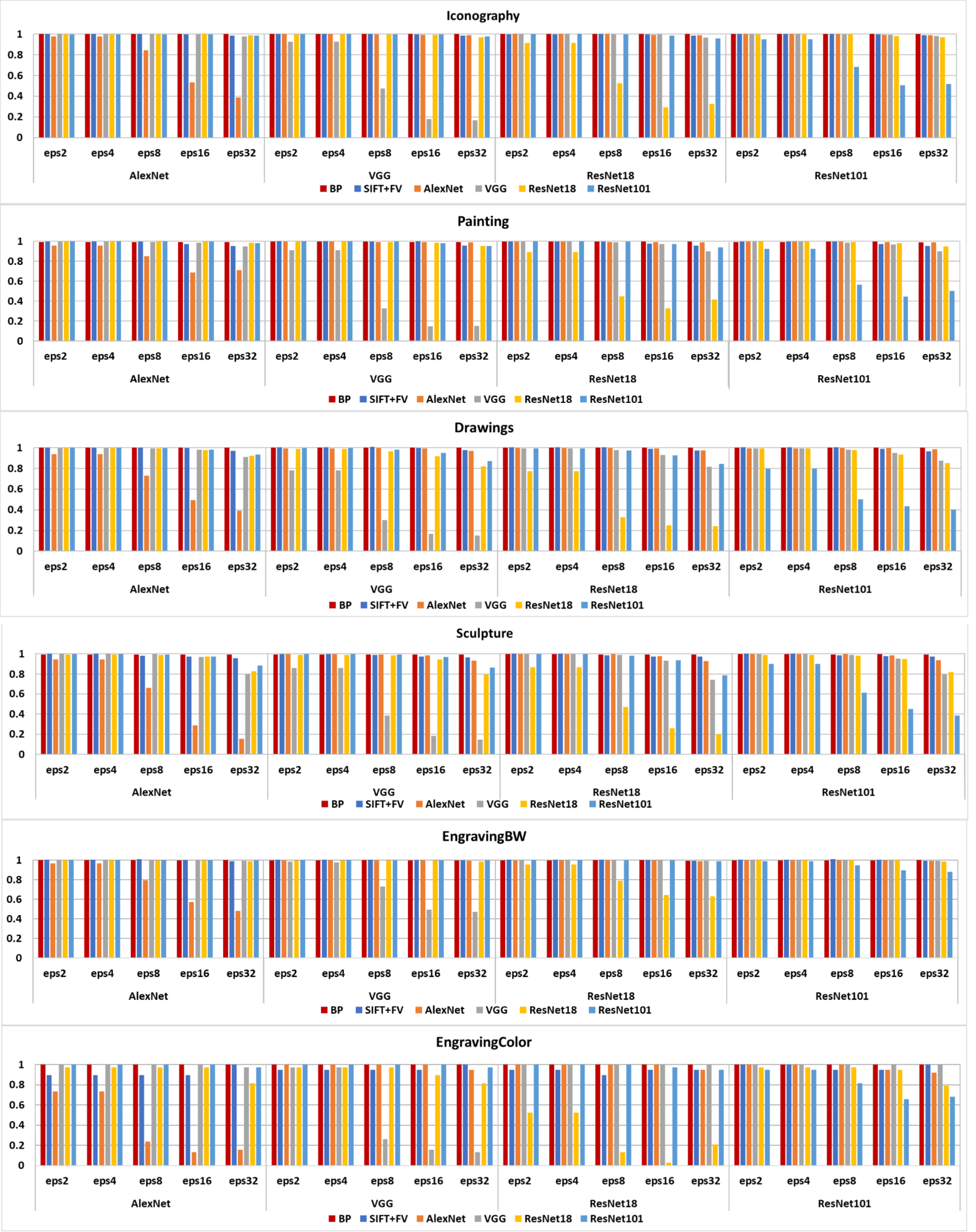}
\caption{Comparative graph of the computed accuracy ratios between adversarial examples and clean images from each method using the validation dataset.}
\label{fig:result}
\end{figure}

Figure~\ref{fig:result} presents the results of Table~\ref{tabkaggle} using the accuracy ratios between adversarial examples and clean images. We observe that the variation of BP is imperceptible in comparison with SIFT+FV and DCNN models. Also, we noted that the performance of DCNNs drastically dropped in almost all classes reaching less than 20\% of its original accuracy when the perturbation matches the network design. In all other cases, the attack reduces the accuracy to about 20\% of the actual performance considering clean images for the classes Sculpture, Engraving BW, and Engraving Color. 

\begin{table}
\caption{Results obtained using training and validation datasets from Kaggle. Each method presents its classification accuracy for training, validation and the AEs using the FGSM computed at $\epsilon={2,4,8,16,32}$.}
\label{tabkaggle}
\begin{center}
\resizebox{\textwidth}{!}
{
\begin{tabular}{|l|l|l|l|l|l|l|l|l|l|l|l|l|l|l|l|l|l|l|l|l|l|l|}
\hline
\multicolumn{23}{|c|}{\textbf{Iconography}}\\ \hline
\multicolumn{3}{|l|}{} & \multicolumn{5}{l|}{AlexNet} & \multicolumn{5}{l|}{VGG} & \multicolumn{5}{l|}{ResNet18
} & \multicolumn{5}{l|}{ResNet101} \\ \hline
& train  & val & $\epsilon2$ & $\epsilon4$ & $\epsilon8$ & $\epsilon16$ & $\epsilon32$ & $\epsilon2$ & $\epsilon4$ & $\epsilon8$ & $\epsilon16$ & $\epsilon32$ & $\epsilon2$ & $\epsilon4$ & $\epsilon8$ & $\epsilon16$ & $\epsilon32$ & $\epsilon2$ & $\epsilon4$ & $\epsilon8$ & $\epsilon16$ & $\epsilon32$  \\ \hline 
BP   & 92.84 & 91.42 & 91.42 & 91.42 & 91.42 & 91.42 & 91.42 & 91.42 & 91.42 & 91.42 & 91.42 & 91.42 & 91.42 & 91.42 & 91.42 & 91.42 & 91.42 & 91.42 & 91.42 & 91.42 & 91.42 & 91.42 \\ \hline
SIFT+FV & 99.92 & 95.91 & 95.91 & 95.91 & 95.91 & 95.59 & 94.26 & 95.83 & 95.83 & 96.07 & 95.52 & 94.57 & 95.75 & 95.75 & 96.07 & 95.52 & 94.34 & 95.99 & 95.99 & 96.07 & 95.67 & 94.73\\ \hline
AlexNet  & 99.61 & 98.66 & 96.3 & 96.3 & 83.24 & 52.56 & 38.39 & 98.51 & 98.51 & 98.43 & 98.03 & 97.64 & 98.58 & 98.58 & 98.66 & 98.03 & 97.4 & 98.51 & 98.51 & 98.51 & 98.03 & 97.48 \\ \hline
VGG  & 100 & 99.21 & 99.29 & 99.29 & 99.06 & 98.82 & 96.85 & 91.9 & 91.9 & 47.05 & 17.7 & 16.76 & 99.21 & 99.21 & 98.98 & 98.74 & 95.83 & 99.21 & 99.21 & 98.98 & 98.35 & 97.32 \\ \hline
ResNet18 & 100 & 98.9 & 98.66 & 98.66 & 98.66 & 98.9 & 97.95 & 98.66 & 98.66 & 98.66 & 98.03 & 95.83 & 90.24 & 90.24 & 52.01 & 29.03 & 32.1 & 98.66 & 98.66 & 98.43 & 97.17 & 95.75 \\ \hline
Resnet101 & 100 & 99.37 & 99.21 & 99.21 & 99.21 & 99.06 & 97.72 & 99.29 & 99.29 & 99.06 & 98.9 & 97.01 & 99.37 & 99.37 & 99.21 & 97.95 & 95.28 & 94.34 & 94.34 & 67.98 & 50.04 & 51.3 \\ 
\hline\hline
\multicolumn{23}{|c|}{\textbf{Painting}}\\ \hline
& train  & val & $\epsilon2$ & $\epsilon4$ & $\epsilon8$ & $\epsilon16$ & $\epsilon32$ & $\epsilon2$ & $\epsilon4$ & $\epsilon8$ & $\epsilon16$ & $\epsilon32$ & $\epsilon2$ & $\epsilon4$ & $\epsilon8$ & $\epsilon16$ & $\epsilon32$ & $\epsilon2$ & $\epsilon4$ & $\epsilon8$ & $\epsilon16$ & $\epsilon32$  \\ \hline 
BP   & 99.68 & 99.04 & 98.25 & 98.25 & 98.48 & 98.41 & 98.48 & 98.78 & 98.8 & 98.64 & 98.33 & 98.41 & 98.8 & 98.8 & 98.56 & 98.64 & 98.56 & 98.41 & 98.41 & 98.56 & 98.8 & 97.69 \\ \hline
SIFT+FV & 99.76 & 92.24 & 92.08 & 92.08 & 92.00 & 89.84 & 87.84 & 92.16 & 92.16 & 92.08 & 90.48 & 88.08 & 91.92 & 91.92 & 91.76 & 90.08 & 88.00 & 92.00 & 92.00 & 91.84 & 89.76 & 87.60\\ \hline
AlexNet  & 98.96 & 97.69 & 93.46 & 93.46 & 83.01 & 66.99 & 69.3 & 97.53 & 97.53 & 97.13 & 96.89 & 96.41 & 97.45 & 97.45 & 96.89 & 96.97 & 96.49 & 97.45 & 97.45 & 97.21 & 97.05 & 96.73\\ \hline
VGG  & 99.92 & 98.17 & 97.93 & 97.93 & 97.53 & 96.73 & 92.82 & 89.31 & 89.31 & 32.14 & 14.27 & 14.91 & 97.69 & 97.69 & 97.05 & 95.45 & 88.28 & 97.69 & 97.69 & 96.81 & 95.14 & 88.12\\ \hline
ResNet18 & 100 & 97.85 & 97.93 & 97.93 & 97.93 & 97.45 & 96.33 & 97.77 & 97.77 & 97.05 & 96.33 & 93.22 & 86.92 & 86.92 & 43.94 & 31.82 & 40.75 & 97.69 & 97.69 & 97.13 & 95.77 & 92.9 \\ \hline
Resnet101 & 100 & 98.56 & 98.72 & 98.72 & 98.48 & 98.17 & 96.65 & 98.64 & 98.64 & 98.25 & 96.49 & 93.86 & 98.72 & 98.72 & 98.17 & 95.85 & 92.58 & 91.15 & 91.15 & 55.42 & 43.94 & 49.68\\ 
\hline \hline
\multicolumn{23}{|c|}{\textbf{Drawings}}\\ \hline
& train  & val & $\epsilon2$ & $\epsilon4$ & $\epsilon8$ & $\epsilon16$ & $\epsilon32$ & $\epsilon2$ & $\epsilon4$ & $\epsilon8$ & $\epsilon16$ & $\epsilon32$ & $\epsilon2$ & $\epsilon4$ & $\epsilon8$ & $\epsilon16$ & $\epsilon32$ & $\epsilon2$ & $\epsilon4$ & $\epsilon8$ & $\epsilon16$ & $\epsilon32$  \\ \hline 
BP   & 96.56 & 90.59 & 90.59 & 90.59 & 90.59 & 90.59 & 90.59 & 90.59 & 90.59 & 90.59 & 90.59 & 90.59 & 90.59 & 90.59 & 90.59 & 90.59 & 90.59 & 90.59 & 90.59 & 90.59 & 90.59 & 90.59\\ \hline
SIFT+FV & 99.87 & 83.84 & 83.84 & 83.84 & 83.97 & 83.46 & 81.30 & 84.22 & 84.22 & 84.48 & 83.59 & 81.93 & 84.22 & 84.22 & 84.22 & 82.95 & 81.68 & 84.10 & 84.10 & 84.35 & 82.95 & 80.79\\ \hline
AlexNet  & 96.44 & 91.35 & 85.75 & 85.75 & 66.79 & 44.91 & 35.62 & 90.84 & 90.84 & 91.22 & 90.59 & 88.55 & 91.09 & 91.09 & 91.09 & 90.59 & 89.06 & 90.71 & 90.71 & 91.09 & 91.09 & 90.08\\ \hline
VGG  & 99.75 & 95.42 & 95.29 & 95.29 & 94.78 & 93.51 & 87.02 & 74.43 & 74.43 & 28.75 & 15.78 & 14.38 & 94.78 & 94.78 & 93.13 & 88.68 & 77.86 & 94.78 & 94.78 & 93.77 & 90.59 & 83.46\\ \hline
ResNet18 & 99.87 & 94.44 & 94.27 & 94.27 & 93.64 & 92.37 & 86.9 & 93.38 & 93.38 & 91.22 & 86.77 & 77.48 & 72.9 & 72.9 & 31.04 & 23.41 & 22.77 & 93.64 & 93.64 & 92.37 & 88.17 & 80.28\\ \hline
Resnet101 & 99.87 & 95.8 & 95.8 & 95.8 & 95.42 & 93.89 & 89.31 & 95.55 & 95.55 & 93.89 & 90.84 & 83.33 & 95.29 & 95.29 & 93.13 & 88.68 & 80.79 & 76.08 & 76.08 & 47.96 & 41.48 & 38.55\\ 
\hline \hline
\multicolumn{23}{|c|}{\textbf{Sculpture}}\\ \hline
& train  & val & $\epsilon2$ & $\epsilon4$ & $\epsilon8$ & $\epsilon16$ & $\epsilon32$ & $\epsilon2$ & $\epsilon4$ & $\epsilon8$ & $\epsilon16$ & $\epsilon32$ & $\epsilon2$ & $\epsilon4$ & $\epsilon8$ & $\epsilon16$ & $\epsilon32$ & $\epsilon2$ & $\epsilon4$ & $\epsilon8$ & $\epsilon16$ & $\epsilon32$  \\ \hline 
BP   & 93.19 & 93.26 & 92.79 & 92.79 & 92.79 & 92.79 & 92.79 & 92.79 & 92.79 & 92.79 & 92.7 & 92.79 & 92.88 & 92.88 & 92.79 & 92.79 & 92.7 & 92.88 & 92.88 & 92.79 & 92.88 & 92.7\\ \hline
SIFT+FV & 99.55 & 87.35 & 87.44 & 87.44 & 85.79 & 85.15 & 83.68 & 87.26 & 87.26 & 86.34 & 85.15 & 84.42 & 87.35 & 87.35 & 85.98 & 84.97 & 85.06 & 87.44 & 87.44 & 85.98 & 85.24 & 85.15\\ \hline
AlexNet  & 99.36 & 95.78 & 90.93 & 90.93 & 63.24 & 27.50 & 14.57 & 95.78 & 95.78 & 95.42 & 94.68 & 89.55 & 95.88 & 95.88 & 95.78 & 94.13 & 89.09 & 95.97 & 95.97 & 96.06 & 94.68 & 90.10 \\ \hline
VGG  & 100 & 97.62 & 98.26 & 98.26 & 97.89 & 94.87 & 78.28 & 84.69 & 84.69 & 37.76 & 17.87 & 14.21 & 98.08 & 98.08 & 97.07 & 91.38 & 72.59 & 97.98 & 97.98 & 96.98 & 93.31 & 78.00\\ \hline
ResNet18 & 100 & 96.88 & 97.25 & 97.25 & 96.88 & 95.05 & 80.66 & 96.88 & 96.88 & 96.15 & 92.39 & 77.54 & 84.88 & 84.88 & 45.92 & 25.30 & 19.07 & 96.70 & 96.70 & 95.69 & 92.58 & 79.65\\ \hline
Resnet101 & 100 & 97.89 & 98.44 & 98.44 & 98.17 & 96.06 & 87.08 & 98.44 & 98.44 & 98.08 & 95.42 & 84.88 & 98.35 & 98.35 & 96.98 & 92.30 & 77.45 & 89.00 & 89.00 & 60.49 & 44.18 & 37.86\\ \hline \hline
\multicolumn{23}{|c|}{\textbf{Engraving BW}}\\ \hline
& train  & val & $\epsilon2$ & $\epsilon4$ & $\epsilon8$ & $\epsilon16$ & $\epsilon32$ & $\epsilon2$ & $\epsilon4$ & $\epsilon8$ & $\epsilon16$ & $\epsilon32$ & $\epsilon2$ & $\epsilon4$ & $\epsilon8$ & $\epsilon16$ & $\epsilon32$ & $\epsilon2$ & $\epsilon4$ & $\epsilon8$ & $\epsilon16$ & $\epsilon32$  \\ \hline 
BP & 89.76 & 92.05 & 92.23 & 92.23 & 92.23 & 91.70 & 91.87 & 91.70 & 91.70 & 92.06 & 91.87 & 91.70 & 91.70 & 91.70 & 92.23 & 92.05 & 91.53 &  91.70 & 91.70 & 91.87 & 91.87 & 92.05 \\ \hline
SIFT+FV & 100 & 93.99 & 94.35 & 94.35 & 94.70 & 94.17 & 92.76 & 94.35 & 94.35 & 94.35 & 94.17 & 93.64 & 94.35 & 94.35 & 94.52 & 94.17 & 93.46 & 94.35 & 94.35 & 94.88 & 94.35 & 93.46\\ \hline
AlexNet  & 99.76 & 99.29 & 96.11 & 96.11 & 78.62 & 56.71 & 47.88 & 99.12 & 99.12 & 99.12 & 98.94 & 98.41 & 99.12 & 99.12 & 99.12 & 98.94 & 98.06 & 99.12 & 99.12 & 99.12 & 98.94 & 98.41\\ \hline
VGG  & 100 & 100 & 99.82 & 99.82 & 99.82 & 99.65 & 99.29 & 98.53 & 97.53 & 73.14 & 49.29 & 47.17 & 99.82 & 99.82 & 99.82 & 99.82 & 99.12 & 99.82 & 99.82 & 99.82 & 99.82 & 99.29\\ \hline
ResNet18 & 100 & 100 & 100 & 100 & 99.82 & 99.82 & 98.94 & 99.82 & 99.82 & 99.82 & 99.65 & 98.23 & 95.58 & 95.58 & 78.98 & 64.49 & 63.07 & 100 & 100 & 100 & 100 & 98.41\\ \hline
Resnet101 & 100 & 100 & 100 & 100 & 100 & 99.82 & 99.47 & 100 & 100 & 99.82 & 99.82 & 99.47 & 100 & 100 & 99.65 & 99.65 & 98.76 & 98.94 & 98.94 & 94.70 & 89.75 & 88.16\\ \hline \hline
\multicolumn{23}{|c|}{\textbf{Engraving Color}}\\ \hline
& train  & val & $\epsilon2$ & $\epsilon4$ & $\epsilon8$ & $\epsilon16$ & $\epsilon32$ & $\epsilon2$ & $\epsilon4$ & $\epsilon8$ & $\epsilon16$ & $\epsilon32$ & $\epsilon2$ & $\epsilon4$ & $\epsilon8$ & $\epsilon16$ & $\epsilon32$ & $\epsilon2$ & $\epsilon4$ & $\epsilon8$ & $\epsilon16$ & $\epsilon32$  \\ \hline 
BP & 98.33 & 97.37 & 97.37 & 97.37 & 97.37 & 97.37 & 97.37 & 97.37 & 97.37 & 97.37 & 97.37 & 97.37 & 97.37 & 97.37 & 97.37 & 97.37 & 97.37 & 97.37 & 97.37 & 97.37 & 97.37 & 97.37 \\ \hline
SIFT+FV & 100 & 50.00 & 44.74 & 44.74 & 44.74 & 44.74 & 50.00 & 47.37 & 47.37 & 47.37 & 47.37 & 50.00 & 47.37 & 47.37 & 44.74 & 47.37 & 47.37 & 50.00 & 50.00 & 47.37 & 47.37 & 50.00\\ \hline
AlexNet  & 100 & 100 & 73.68 & 73.68 & 23.68 & 13.16 & 15.79 & 100 & 100 & 100 & 100 & 94.74 & 100 & 100 & 100 & 100 & 94.74 & 100 & 100 & 100 & 94.74 & 92.11\\ \hline
VGG  & 100 & 100 & 100 & 100 & 100 & 100 & 97.37 & 97.37 & 97.37 & 26.32 & 15.79 & 13.16 & 100 & 100 & 100 & 100 & 100 & 100 & 100 & 100 & 100 & 100\\ \hline
ResNet18 & 95.00 & 100 & 97.37 & 97.37 & 97.37 & 97.37 & 81.58 & 97.37 & 97.37 & 97.37 & 89.47 & 81.58 & 52.63 & 52.63 & 13.16 & 02.63 & 21.05 & 97.37 & 97.37 & 97.37 & 94.74 & 78.95\\ \hline
Resnet101 & 100 & 100 & 100 & 100 & 100 & 100 & 97.37 & 100 & 100 & 100 & 100 & 97.37 & 100 & 100 & 100 & 97.37 & 94.74 & 94.74 & 94.74 & 81.58 & 65.79 & 68.42\\ 
\hline
\end{tabular}
}
 \end{center}
\end{table}

\begin{table}
\caption{Results of using the SVM hyperparameters optimizer method from Matlab and the crossvalidation function to verify overfitting on SIFT+FV.}
\label{crossvalidation}
\begin{center}
\begin{tabular}{|l|l|l|l|l|}
\hline
& \multicolumn{2}{|l|}{optimizer} & \multicolumn{2}{|l|}{cross-validation} \\ \hline
\textbf{SIFT+FV} & train  & val & mean train & mean val \\ \hline
Iconography & 100 & 95.28 & 99.28 & 95.28  \\ \hline
Painting & 99.76 & 92.72 & 98.84 & 92.83   \\ \hline
Drawings  & 100 & 83.84 & 98.28 & 83.44   \\ \hline
Sculpture  & 100 & 86.71 & 98.63 & 86.48 \\ \hline
Engraving Bw  & 100 & 93.64 & 99.32 & 93.87   \\ \hline
Engraving Color  & 100 & 50.00 & 92.00 & 47.11 \\ \hline
\hline
\end{tabular}
 \end{center}
\end{table}

\begin{figure}
\centering
\includegraphics[width=0.7\textwidth]{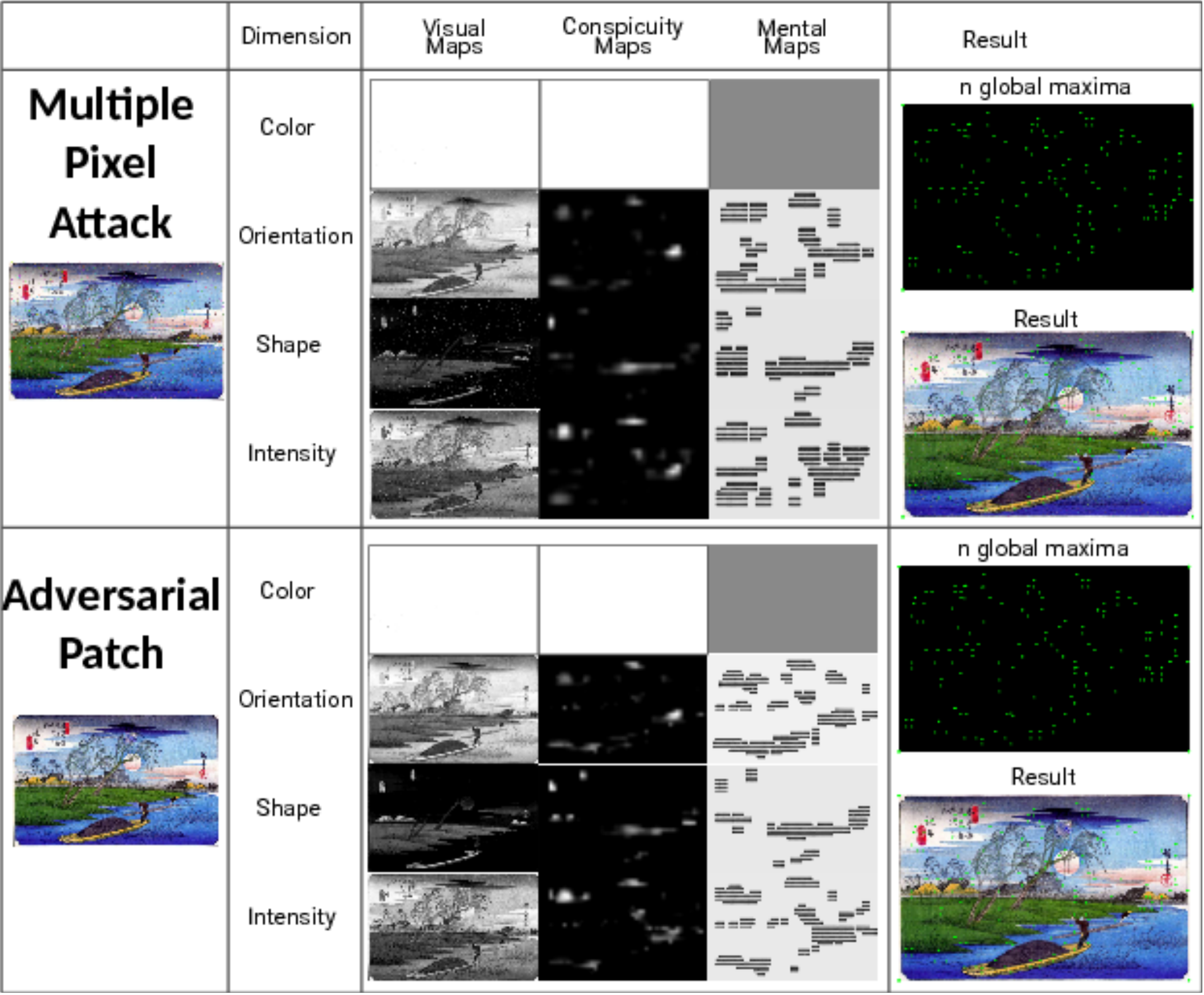}
\caption{Maps generated in each phase of the AVC, extracted from an AE of the multiple pixel attack and the image with the adversarial patch. Note that despite the attack, the generated maps do not change much with the original one.}
\label{fig:dimensionOne}
\end{figure}

\begin{figure}
\centering
\includegraphics[width=0.95\textwidth]{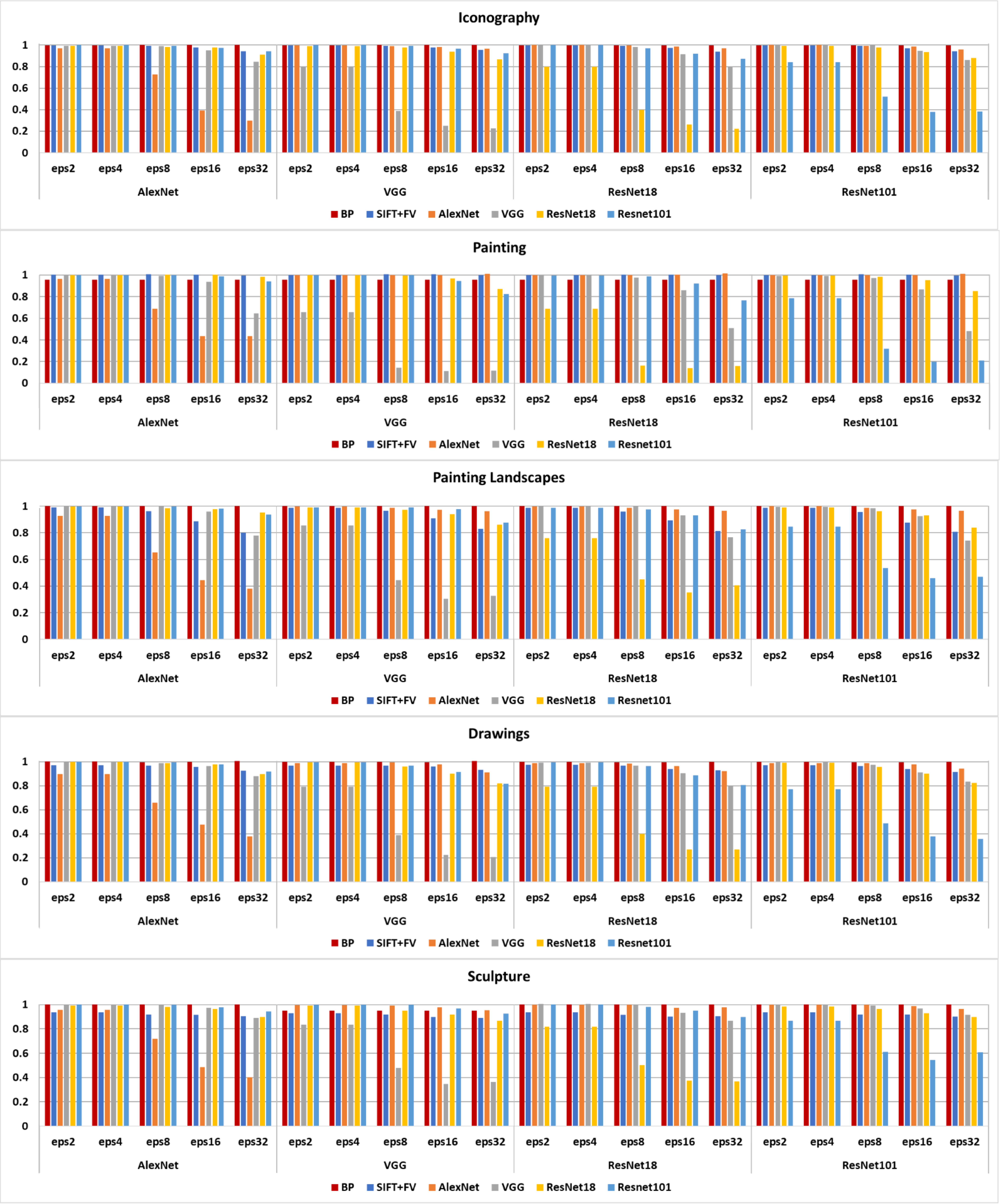}

\caption{Comparative graph of the computed accuracy ratios between adversarial examples and clean images from each method using Iconography, Painting, Painting Landscapes, Drawings, and Sculpture classes from the testing dataset.}
\label{fig:resultwiki1}
\end{figure}

\begin{figure}
\centering
\includegraphics[width=0.95\textwidth]{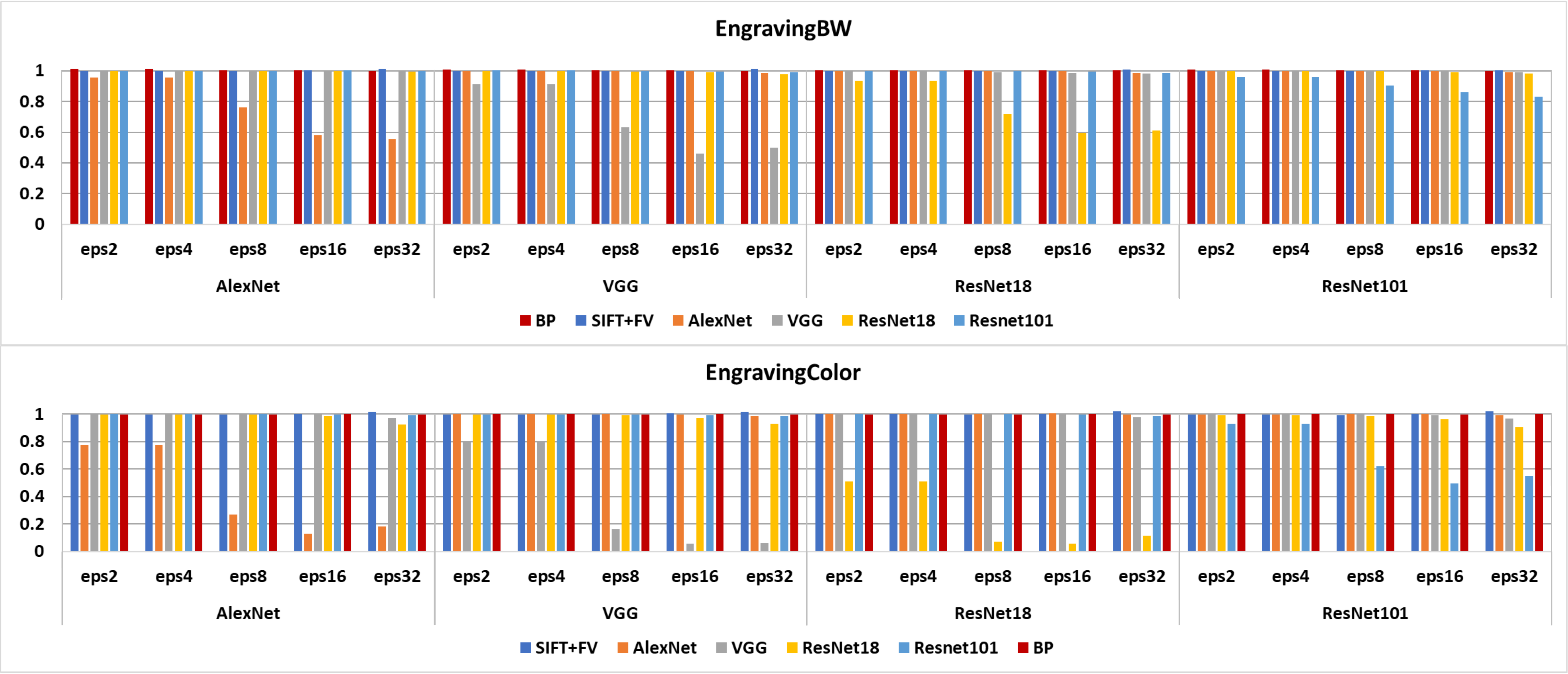}

\caption{Comparative graph of the computed accuracy ratios between adversarial examples and clean images from each method using Engraving BW and Engraving Color classes from the testing dataset.}
\label{fig:resultwiki2}
\end{figure}

The testing stage exhibited an even worse behavior compared with the validation dataset for the DCNN and SIFT+FV. The drop in the transferability performance was higher when the scaling factor $\epsilon$ becomes larger. Table~\ref{tabwiki} shows that the accuracy was compromised in all DCNN models for three classes: Painting, Drawings, and Engraving Color. For example, the worst-case is Engraving Color, where AlexNet fell to 17.22\% from a clean score of 94.72\%, VGG and ResNet18 diminished their performance to almost 5\% of accuracy after scoring 99\% and 96\%, respectively, and ResNet101 achieves 49\%, which was the less affected in accuracy. Moreover, the experimental results in Table~\ref{tabwiki} provide the FGSM transferability in DCNN models. Notice that the effect on $\epsilon=32$ reaches the more significant changes. Also, the test showed the poor performance of SIFT+FV considering clean images. In four out of seven classes (Painting Landscapes, Drawings, Sculpture, and Engraving color), the accuracy is way below to compete with DCNNs. Additionally, SIFT+FV was affected by AEs in Iconography, Painting Landscapes, and Sculpture, where approximately 10\% of its original score reduced the performance. Finally, BP demonstrated high quality and steady results keeping its scores from clean images after AEs with minimal to zero changes for all classes.

Additionally, it is noticeable that, as opposed to SIFT+FV, BP reaches comparable results to DCNNs' scores.  Besides, we present in Figures~\ref{fig:resultwiki1}-\ref{fig:resultwiki2} the ratio of accuracy on AEs for the testing classes. We observed a very similar behavior, at least for BP, whose rate for all experiments remains almost one. We see a drastic drop in DCNN models' performance when the perturbation matches the network's architecture and influences AEs' transferability to other DCNN models and SIFT+FV.

\begin{table}
\caption{Results obtained using the test dataset from Wikiart. Each method presents its classification accuracy for testing and the AEs using the FGSM computed at $\epsilon={2,4,8,16,32}$.}
\label{tabwiki}
\begin{center}
\resizebox{\textwidth}{!}
{
\begin{tabular}{|l|l||l|l|l|l|l||l|l|l|l|l||l|l|l|l|l||l|l|l|l|l|}
\hline
\multicolumn{22}{|c|}{\textbf{Iconography}}\\ \hline
\multicolumn{2}{|l||}{} & \multicolumn{5}{l||}{AlexNet} & \multicolumn{5}{l||}{VGG} & \multicolumn{5}{l||}{ResNet18
} & \multicolumn{5}{l||}{ResNet101} \\ \hline
& test & $\epsilon2$ & $\epsilon4$ & $\epsilon8$ & $\epsilon16$ & $\epsilon32$ & $\epsilon2$ & $\epsilon4$ & $\epsilon8$ & $\epsilon16$ & $\epsilon32$ & $\epsilon2$ & $\epsilon4$ & $\epsilon8$ & $\epsilon16$ & $\epsilon32$ & $\epsilon2$ & $\epsilon4$ & $\epsilon8$ & $\epsilon16$ & $\epsilon32$  \\ \hline 
BP & 91.74 & 91.66 & 91.66 & 91.82 & 91.74 & 91.74 & 91.66 & 91.66 & 91.74 & 91.74 & 91.74 & 91.66 & 91.66 & 91.66 & 91.58 & 91.5 & 91.58 & 91.58 & 91.58 & 91.58 & 91.58\\ \hline
SIFT+FV & 86.16 & 85.54 & 85.54 & 84.71 & 83.26 & 77.69 & 85.54 & 85.54 & 84.92 & 83.47 & 77.48 & 85.95 & 85.95 & 84.71 & 83.06 & 76.24 & 86.16 & 86.16 & 84.71 & 83.06 & 75.62\\ \hline
AlexNet  & 96.07 & 93.39 & 93.39 & 70.04 & 37.4 & 28.72 & 95.87 & 95.87 & 95.04 & 94.42 & 92.98 & 96.07 & 96.07 & 95.87 & 94.83 & 93.18 & 96.07 & 96.07 & 95.45 & 94.63 & 92.15\\ \hline
VGG  & 95.87 & 95.45 & 95.45 & 94.83 & 91.32 & 80.99 & 76.65 & 76.65 & 36.98 & 23.97 & 21.69 & 95.66 & 95.66 & 94.21 & 87.81 & 76.86 & 95.87 & 95.87 & 95.87 & 90.91 & 82.44\\ \hline
ResNet18 & 96.49 & 95.87 & 95.87 & 94.83 & 94.21 & 87.81 & 95.66 & 95.66 & 94.42 & 90.5 & 83.88 & 76.86 & 76.86 & 38.64 & 25.21 & 21.49 & 96.07 & 96.07 & 94.21 & 90.29 & 85.12\\ \hline
Resnet101 & 95.25 & 95.25 & 95.25 & 94.83 & 92.77 & 89.88 & 95.45 & 95.45 & 94.63 & 91.94 & 88.02 & 95.45 & 95.45 & 92.56 & 87.6 & 83.26 & 79.96 & 79.96 & 49.38 & 36.16 & 36.36\\ 
\hline\hline
\multicolumn{22}{|c|}{\textbf{Painting}}\\ \hline
& test & $\epsilon2$ & $\epsilon4$ & $\epsilon8$ & $\epsilon16$ & $\epsilon32$ & $\epsilon2$ & $\epsilon4$ & $\epsilon8$ & $\epsilon16$ & $\epsilon32$ & $\epsilon2$ & $\epsilon4$ & $\epsilon8$ & $\epsilon16$ & $\epsilon32$ & $\epsilon2$ & $\epsilon4$ & $\epsilon8$ & $\epsilon16$ & $\epsilon32$  \\ \hline
BP & 100 & 95.65 & 95.65 & 95.65 & 95.65 & 95.65 & 95.65 & 95.65 & 95.65 & 95.65 & 95.65 & 95.65 & 95.65 & 95.65 & 95.65 & 95.65 & 95.65 & 95.65 & 95.65 & 95.65 & 95.65\\ \hline
SIFT+FV & 94.83 & 94.83 & 94.83 & 94.70 & 94.57 & 93.63 & 94.92 & 94.92 & 94.88 & 94.57 & 93.63 & 94.88 & 94.88 & 94.79 & 94.44 & 93.28 & 94.88 & 94.88 & 94.70 & 94.32 & 93.20\\ \hline
AlexNet  & 94.06 & 90.57 & 90.57 & 64.64 & 41.04 & 41.00 & 94.10 & 94.10 & 93.90 & 94.01 & 94.92 & 94.10 & 94.10 & 94.06 & 94.32 & 95.35 & 94.10 & 94.10 & 94.06 & 94.06 & 95.00\\ \hline
VGG  & 93.37 & 93.28 & 93.28 & 92.64 & 87.47 & 60.12 & 61.15 & 61.15 & 13.14 & 10.42 & 10.68 & 92.89 & 92.89 & 91.17 & 80.10 & 47.55 & 92.59 & 92.59 & 90.78 & 81.05 & 44.96\\ \hline
ResNet18 & 94.23 & 94.19 & 94.19 & 94.40 & 94.40 & 92.64 & 94.01 & 94.01 & 93.63 & 91.30 & 81.91 & 64.86 & 64.86 & 15.25 & 13.01 & 15.07 & 93.80 & 93.80 & 92.72 & 89.84 & 80.19\\ \hline
Resnet101 & 95.91 & 95.82 & 95.82 & 95.78 & 94.62 & 90.09 & 95.82 & 95.82 & 95.69 & 90.44 & 79.03 & 95.61 & 95.61 & 94.66 & 88.33 & 73.47 & 75.24 & 75.24 & 30.62 & 19.04 & 19.98\\ 
\hline\hline
\multicolumn{22}{|c|}{\textbf{Painting Landscapes}}\\ \hline
& test & $\epsilon2$ & $\epsilon4$ & $\epsilon8$ & $\epsilon16$ & $\epsilon32$ & $\epsilon2$ & $\epsilon4$ & $\epsilon8$ & $\epsilon16$ & $\epsilon32$ & $\epsilon2$ & $\epsilon4$ & $\epsilon8$ & $\epsilon16$ & $\epsilon32$ & $\epsilon2$ & $\epsilon4$ & $\epsilon8$ & $\epsilon16$ & $\epsilon32$  \\ \hline 
BP & 100 & 100 & 100 & 100 & 100 & 100 & 100 & 100 & 100 & 100 & 100 & 100 & 100 & 100 & 100 & 100 & 100 & 100 & 100 & 100 & 100\\ \hline
SIFT+FV & 75.34 & 75.07 & 75.07 & 72.90 & 70.46 & 62.6 & 75.61 & 75.61 & 73.71 & 70.46 & 62.60 & 75.34 & 75.34 & 73.17 & 68.83 & 60.70 & 75.34 & 75.34 & 72.90 & 68.29 & 59.62\\ \hline
AlexNet  & 93.77 & 86.99 & 86.99 & 61.25 & 41.46 & 35.77 & 93.50 & 93.50 & 92.68 & 91.06 & 90.24 & 93.50 & 93.50 & 92.68 & 91.60 & 90.51 & 93.77 & 93.77 & 92.68 & 91.33 & 90.51\\ \hline
VGG  & 94.58 & 94.58 & 94.58 & 94.31 & 90.79 & 73.71 & 80.76 & 80.76 & 42.01 & 28.73 & 30.89 & 94.31 & 94.31 & 94.31 & 88.08 & 72.63 & 94.04 & 94.04 & 93.22 & 87.53 & 70.19\\ \hline
ResNet18 & 95.12 & 94.85 & 94.85 & 93.77 & 92.95 & 90.79 & 94.31 & 94.31 & 92.41 & 89.43 & 81.84 & 72.36 & 72.36 & 42.82 & 33.60 & 38.48 & 94.31 & 94.31 & 91.6 & 88.62 & 79.95\\ \hline
Resnet101 & 95.39 & 95.12 & 95.12 & 95.12 & 93.77 & 89.43 & 94.58 & 94.58 & 94.58 & 93.50 & 83.74 & 94.31 & 94.31 & 92.95 & 88.89 & 78.86 & 80.76 & 80.76 & 50.96 & 43.90 & 44.72\\ 
\hline\hline
\multicolumn{22}{|c|}{\textbf{Drawings}}\\ \hline
& test & $\epsilon2$ & $\epsilon4$ & $\epsilon8$ & $\epsilon16$ & $\epsilon32$ & $\epsilon2$ & $\epsilon4$ & $\epsilon8$ & $\epsilon16$ & $\epsilon32$ & $\epsilon2$ & $\epsilon4$ & $\epsilon8$ & $\epsilon16$ & $\epsilon32$ & $\epsilon2$ & $\epsilon4$ & $\epsilon8$ & $\epsilon16$ & $\epsilon32$  \\ \hline 
BP & 94.05 & 94.28 & 94.28 & 93.59 & 93.81 & 94.5 & 93.82 & 93.82 & 94.05 & 93.59 & 94.73 & 93.81 & 93.81 & 93.59 & 93.59 & 94.05 & 93.81 & 93.81 & 93.59 & 93.59 & 93.81\\ \hline
SIFT+FV & 73.61 & 68.42 & 68.42 & 66.36 & 62.7 & 56.52 & 68.42 & 68.42 & 67.51 & 62.7 & 57.89 & 68.42 & 68.42 & 66.82 & 62.01 & 56.75 & 68.19 & 68.19 & 67.28 & 62.01 & 56.75\\ \hline
AlexNet  & 86.73 & 77.8 & 77.8 & 57.21 & 41.19 & 32.72 & 85.81 & 85.81 & 86.27 & 84.67 & 79.18 & 85.81 & 85.81 & 85.35 & 83.75 & 80.09 & 85.81 & 85.81 & 85.58 & 84.67 & 81.69\\ \hline
VGG  & 91.99 & 91.99 & 91.99 & 90.89 & 88.79 & 80.78 & 72.77 & 72.77 & 35.7 & 20.59 & 18.99 & 91.3 & 91.3 & 89.02 & 83.3 & 73.91 & 91.53 & 91.53 & 89.7 & 83.98 & 76.89\\ \hline
ResNet18 & 90.85 & 90.85 & 90.85 & 89.7 & 88.79 & 81.46 & 90.39 & 90.39 & 87.41 & 81.69 & 74.37 & 71.85 & 71.85 & 36.16 & 24.49 & 24.49 & 90.16 & 90.16 & 86.96 & 81.92 & 74.83\\ \hline
Resnet101 & 93.59 & 93.36 & 93.36 & 93.14 & 91.53 & 85.81 & 93.14 & 93.14 & 90.62 & 85.58 & 76.43 & 93.14 & 93.14 & 90.16 & 83.07 & 75.29 & 72.27 & 72.27 & 45.54 & 35.24 & 33.41\\ 
\hline\hline
\multicolumn{22}{|c|}{\textbf{Sculpture}}\\ \hline
& test & $\epsilon2$ & $\epsilon4$ & $\epsilon8$ & $\epsilon16$ & $\epsilon32$ & $\epsilon2$ & $\epsilon4$ & $\epsilon8$ & $\epsilon16$ & $\epsilon32$ & $\epsilon2$ & $\epsilon4$ & $\epsilon8$ & $\epsilon16$ & $\epsilon32$ & $\epsilon2$ & $\epsilon4$ & $\epsilon8$ & $\epsilon16$ & $\epsilon32$  \\ \hline 
BP & 90.54 & 90.83 & 90.83 & 90.83 & 90.83 & 90.83 & 85.96 & 85.96 & 85.96 & 85.96 & 85.96 & 90.83 & 90.83 & 90.83 & 90.83 & 90.83 & 90.83 & 90.83 & 90.83 & 90.83 & 90.83 \\ \hline
SIFT+FV & 60.47 & 52.80 & 52.80 & 52.80 & 53.10 & 51.62 & 53.39 & 53.39 & 53.10 & 52.51 & 52.51 & 52.80 & 52.80 & 52.51 & 52.51 & 51.33 & 53.39 & 53.39 & 52.51 & 52.51 & 50.74\\ \hline
AlexNet  & 91.45 & 87.61 & 87.61 & 65.49 & 44.25 & 36.87 & 91.15 & 91.15 & 90.56 & 89.38 & 87.32 & 91.45 & 91.45 & 91.45 & 89.09 & 89.38 & 91.45 & 91.45 & 91.45 & 90.27 & 88.20\\ \hline
VGG  & 94.69 & 94.99 & 94.99 & 94.99 & 92.33 & 84.37 & 79.06 & 79.06 & 45.43 & 32.74 & 34.51 & 95.28 & 95.28 & 94.10 & 88.20 & 82.01 & 94.69 & 94.69 & 93.81 & 91.74 & 86.73\\ \hline
ResNet18 & 92.63 & 91.74 & 91.74 & 90.86 & 89.38 & 83.19 & 91.74 & 91.74 & 87.91 & 84.96 & 80.24 & 75.81 & 75.81 & 46.61 & 34.81 & 33.92 & 91.15 & 91.15 & 89.38 & 86.14 & 83.19\\ \hline
Resnet101 & 92.92 & 93.22 & 93.22 & 92.63 & 90.86 & 87.61 & 92.92 & 92.92 & 92.92 & 89.97 & 86.14 & 93.22 & 93.22 & 91.15 & 88.20 & 83.48 & 80.53 & 80.53 & 56.64 & 50.44 & 56.34\\ 
\hline\hline
\multicolumn{22}{|c|}{\textbf{Engraving BW}}\\ \hline
& test & $\epsilon2$ & $\epsilon4$ & $\epsilon8$ & $\epsilon16$ & $\epsilon32$ & $\epsilon2$ & $\epsilon4$ & $\epsilon8$ & $\epsilon16$ & $\epsilon32$ & $\epsilon2$ & $\epsilon4$ & $\epsilon8$ & $\epsilon16$ & $\epsilon32$ & $\epsilon2$ & $\epsilon4$ & $\epsilon8$ & $\epsilon16$ & $\epsilon32$  \\ \hline 
BP & 91.55 & 92.64 & 92.64 & 91.97 & 91.72 & 91.63 & 92.30 & 92.30 & 92.05 & 92.05 & 91.63 & 91.97 & 91.97 & 91.80 & 91.97 & 91.80 & 92.13 & 92.13 & 91.97 & 91.80 & 91.63 \\ \hline
SIFT+FV & 89.79 & 89.79 & 89.79 & 89.71 & 89.87 & 90.96 & 89.79 & 89.79 & 89.79 & 89.37 & 90.96 & 89.87 & 89.87 & 89.62 & 89.54 & 90.88 & 89.96 & 89.96 & 89.46 & 89.54 & 90.96\\ \hline
AlexNet  & 98.58 & 94.06 & 94.06 & 75.06 & 57.32 & 54.64 & 98.66 & 98.66 & 98.66 & 98.49 & 97.32 & 98.66 & 98.66 & 98.66 & 98.33 & 97.15 & 98.66 & 98.66 & 98.66 & 98.58 & 97.49\\ \hline
VGG  & 99.58 & 99.83 & 99.83 & 99.67 & 99.50 & 99.16 & 91.05 & 91.05 & 62.85 & 45.94 & 49.87 & 99.58 & 99.58 & 98.74 & 98.41 & 97.91 & 99.58 & 99.58 & 99.25 & 99.00 & 98.83\\ \hline
ResNet18 & 99.83 & 99.92 & 99.92 & 99.83 & 99.67 & 99.16 & 99.75 & 99.75 & 99.41 & 98.83 & 97.49 & 93.22 & 93.22 & 71.55 & 59.41 & 61.09 & 99.83 & 99.83 & 99.67 & 98.91 & 97.82\\ \hline
Resnet101 & 99.67 & 99.75 & 99.75 & 99.75 & 99.83 & 99.75 & 99.83 & 99.83 & 99.50 & 99.25 & 98.74 & 99.67 & 99.67 & 99.50 & 99.08 & 98.16 & 95.90 & 95.90 & 90.13 & 85.77 & 83.01\\ 
\hline\hline
\multicolumn{22}{|c|}{\textbf{Engraving Color}}\\ \hline
& test & $\epsilon2$ & $\epsilon4$ & $\epsilon8$ & $\epsilon16$ & $\epsilon32$ & $\epsilon2$ & $\epsilon4$ & $\epsilon8$ & $\epsilon16$ & $\epsilon32$ & $\epsilon2$ & $\epsilon4$ & $\epsilon8$ & $\epsilon16$ & $\epsilon32$ & $\epsilon2$ & $\epsilon4$ & $\epsilon8$ & $\epsilon16$ & $\epsilon32$  \\ \hline 
BP & 89.92 & 89.68 & 89.68 & 89.74 & 89.86 & 89.80 & 89.92 & 89.92 & 89.74 & 89.86 & 89.62 & 89.68  & 89.68 & 89.74 & 89.98 & 89.80 & 89.92 & 89.92 & 89.86 & 89.50 & 90.16 \\ \hline
SIFT+FV & 66.95 & 66.77 & 66.77 & 66.59 & 66.89 & 68.09 & 66.83 & 66.83 & 66.59 & 67.19 & 68.09 & 66.89 & 66.65 & 66.95 & 68.33 & 66.71 & 66.71 & 66.53 & 66.53 & 66.95 & 66.95\\ \hline
AlexNet  & 94.72 & 73.55 & 73.55 & 25.49 & 12.30 & 17.22 & 94.78 & 94.78 & 94.90 & 94.48 & 93.64 & 94.72 & 94.72 & 94.66 & 95.14 & 94.24 & 94.54 & 94.54 & 95.02 & 94.66 & 94.00\\ \hline
VGG  & 99.40 & 99.46 & 99.46 & 99.46 & 99.28 & 96.52 & 79.90 & 79.90 & 16.02 & 05.46 & 06.06 & 99.52 & 99.52 & 99.22 & 99.10 & 97.18 & 99.40 & 99.40 & 99.10 & 98.50 & 95.98\\ \hline
ResNet18 & 96.40 & 95.98 & 95.98 & 96.16 & 95.02 & 89.14 & 95.92 & 95.92 & 95.50 & 93.88 & 89.50 & 49.13 & 49.13 & 06.84 & 05.58 & 10.74 & 95.62 & 95.62 & 95.02 & 92.68 & 86.98\\ \hline
Resnet101 & 99.88 & 99.76 & 99.76 & 99.76 & 99.52 & 98.92 & 99.70 & 99.70 & 99.70 & 99.22 & 98.44 & 99.82 & 99.82 & 99.76 & 99.40 & 98.56 & 92.86 & 92.86 & 61.91 & 49.19 & 54.53\\ 
\hline
\end{tabular}
}
\end{center}
\end{table}

\subsubsection{Multiple Pixel Attack}

The multiple pixel attack experiment came along with the analysis that one pixel does not perturb high-resolution images to change the model's prediction. We experiment with modifying one pixel to fool the models over 100 selected images, and the results indicate no score changes. Thus, we experimentally found that when 8000-10,000 pixels, DCNN models have a massive amount of change in their prediction, so we set a second experiment with a 10,000 pixel attack. We present in Table~\ref{tabonepixel} the number of images that change its forecast with the success rate and the mean posterior probability of these new predictions in the confidence row.

We observed that DCNN changes by a considerable amount of their predictions with high confidence by modifying multiple pixels. SIFT+FV was also misled in five out of seven classes achieving the same number of images as DCNN models with lower confidence. In this way, only two categories resisted the attack. On the contrary, BP was robust to this attack having four out of seven classes without changes and the rest with a maximum error of 4\%. Notice that the amount of pixels modified in this experiment fails the motivation of AA in which the perturbation should be unnoticeable to human vision. Therefore, BP was robust to this perturbation. We illustrate as an example; BP generated maps using a multiple pixel attack in Figure \ref{fig:dimensionOne}. Additionally, we report the mean processing time in seconds (see Table~\ref{tabonepixel}), which makes this attack unfeasible to perform in real-time applications.

\begin{table}
\caption{Results from the experiment of computing the multiple pixel attack with $d=10,000$ on 100 random images from the testing dataset. The original accuracy refers to the score of the clean images. Success rate means the percentage of images that change the prediction with a mean Confidence value of the posterior probabilities over the new predicted classes.}
\label{tabonepixel}
\begin{center}
\resizebox{\textwidth}{!}{
\begin{tabular}{|l|l|l|l|l|l|l|}
\hline
\textbf{Iconography} & BP  & SIFT+FV & AlexNet & VGG & ResNet18 & ResNet101 \\ \hline
Original Acc. & 92.00 & 88.00 & 96.00 & 94.00 & 96.00 & 92.00 \\ \hline
Success rate & 0.00 & 32.00 & 32.00 & 44.00 & 46.00 & 42.00 \\ \hline
Confidence  & NA & 64.96 & 85.09 & 85.72 & 76.34 & 77.61 \\ \hline
Time (seconds)  & 94.22 & 301.21 & 138.51 & 147.72 & 152.37 & 237.73  \\ \hline\hline
\textbf{Painting} & BP  & SIFT+FV & AlexNet & VGG & ResNet18 & ResNet101 \\ \hline
Original Acc. & 100 & 78.00 & 94.00 & 90.00 & 92.00 & 94.00 \\ \hline
Success rate & 2.00 & 0.00 & 54.00 & 60.00 & 64.00 & 64.00 \\ \hline
Confidence  & 51.83 & NA & 78.11 & 97.34 & 99.37 & 98.06 \\ \hline
Time (seconds)  & 90.16 & 598.12 & 119.78 & 122.59 & 111.14 & 242.58  \\ \hline\hline
\textbf{Painting Landscapes} & BP  & SIFT+FV & AlexNet & VGG & ResNet18 & ResNet101 \\ \hline
Original Acc. & 100 & 78.00 & 88.00 & 88.00 & 92.00 & 92.00 \\ \hline
Success rate & 2.00 & 40.00 & 54.00 & 60.00 & 64.00 & 66.00 \\ \hline
Confidence  & 54.06 & 62.04 & 75.70 & 97.25 & 99.26 & 97.37 \\ \hline
Time (seconds)  & 98.83 & 585.69 & 141.85 & 163.51 & 143.62 & 205.53  \\ \hline\hline
\textbf{Drawings} & BP  & SIFT+FV & AlexNet & VGG & ResNet18 & ResNet101 \\ \hline
Original Acc. & 88.00 & 70.00 & 80.00 & 90.00 & 86.00 & 92.00 \\ \hline
Success rate & 0.00 & 38.00 & 68.00 & 68.00 & 74.00 & 78.00 \\ \hline
Confidence  & NA & 66.53 & 83.91 & 91.94 & 95.11 & 94.24 \\ \hline
Time (seconds)  & 118.85 & 462.92 & 110.18 & 111.48 & 128.07 & 220.69  \\ \hline\hline
\textbf{Sculpture} & BP  & SIFT+FV & AlexNet & VGG & ResNet18 & ResNet101 \\ \hline
Original Acc. & 86.00 & 62.00 & 88.00 & 98.00 & 96.00 & 96.00 \\ \hline
Success rate & 4.00 & 60.00 & 62.00 & 54.00 & 56.00 & 54.00 \\ \hline
Confidence  & 58.14 & 67.61 & 92.65 & 98.60 & 97.45 & 96.93 \\ \hline
Time (seconds)  & 71.20 & 601.53 & 121.22 & 130.06 & 137.16 & 181.14  \\ \hline\hline
\textbf{Engraving BW} & BP  & SIFT+FV & AlexNet & VGG & ResNet18 & ResNet101 \\ \hline
Original Acc. & 94.00 & 94.00 & 100 & 100 & 100 & 100 \\ \hline
Success rate & 0.00 & 0.00 & 40.00 & 50.00 & 32.00 & 20.00 \\ \hline
Confidence  & NA & NA & 77.63 & 68.07 & 71.86 & 61.25 \\ \hline
Time (seconds)  & 88.71 & 599.41 & 148.90 & 169.56 & 152.11 & 177.61  \\ \hline \hline
\textbf{Engraving Color} & BP  & SIFT+FV & AlexNet & VGG & ResNet18 & ResNet101 \\ \hline
Original Acc. & 94.00 & 74.00 & 98.00 & 100 & 92.00 & 100 \\ \hline
Success rate & 0.00 & 60.00 & 40.00 & 50.00 & 46.00 & 22.00 \\ \hline
Confidence  & NA & 55.98 & 73.80 & 66.15 & 62.96 & 65.31 \\ \hline
Time (seconds)  & 87.01 & 600.82 & 150.70 & 174.51 & 154.52 & 186.13  \\ \hline
 
\hline
\end{tabular}
}
\end{center}
\end{table}

\subsubsection{Adversarial Patch}

We present the results of the adversarial patch in Table~\ref{tabpatch}. This experiment analyzes the change in the model's predictions by adding the trained patches from DCNN models using 100 images from each class in a random location and orientation. The results from Table~\ref{tabpatch} show that these patches affect in a significant manner DCNN models in most experiments. Also, we discovered that the patches could be transferable to other DCNNs. 

The painting landscapes experiment showed the worst-case scenario for DCNN models, on which we observed a considerable transferability effect between the models. We observed that VGG, ResNet18, and ResNet101 were affected by all the patches. DCNN models dropped its performance to approximately half of its original accuracy and, in some cases, is less to 50\%. ResNet18 was fooled in all images using its trained patch. All other classes did not show a similar behavior; the patches can fool DCNN models. In contrast, SIFT+FV and BP demonstrated a robust control over the adversarial patches, showing almost an unchangeable performance. Figure \ref{fig:dimensionOne} illustrates the BP generated maps using an image with the adversarial patch. 


\begin{table}
\caption{Results obtained using the adversarial patch. Each column presents the score obtained for the original 100 images per class and the AEs when adding the adversarial patch.}
\label{tabpatch}
\begin{center}
\resizebox{0.7\textwidth}{!}
{
\begin{tabular}{|l|l|l|l|l|l|l|}
\hline
\textbf{Iconography} & Original Acc. & AlexNet Patch & VGG Patch & ResNet18 Patch & ResNet101 Patch\\ \hline
BP & 99.00 & 99.00 & 99.00 & 99.00 & 99.00 \\ \hline
SIFT+FV & 92.00 & 89.00 & 93.00 & 93.00 & 92.00 \\ \hline
AlexNet & 98.00 & 74.00 & 97.00 & 97.00 & 98.00 \\ \hline
VGG & 94.00 & 91.00 & 45.00 & 82.00 & 81.00 \\ \hline
ResNet18  & 94.00 & 87.00 & 90.00 & 58.00 & 90.00 \\ \hline
ResNet101  & 93.00 & 87.00 & 87.00 & 78.00 & 70.00 \\ \hline
\hline
\textbf{Painting} & Original Acc. & AlexNet Patch & VGG Patch & ResNet18 Patch & ResNet101 Patch\\ \hline
BP & 100.00 & 100.00 & 100.00 & 99.00 & 100.00 \\ \hline
SIFT+FV & 97.00 & 98.00 & 97.00 & 98.00 & 96.00 \\ \hline
AlexNet & 96.00 & 54.00 & 94.00 & 94.00 & 94.00 \\ \hline
VGG & 92.00 & 71.00 & 48.00 & 73.00 & 61.00 \\ \hline
ResNet18  & 94.00 & 67.00 & 76.00 & 23.00 & 43.00 \\ \hline
ResNet101  & 97.00 & 72.00 & 72.00 & 69.00 & 56.00 \\ \hline
\hline
\textbf{Painting Land.} & Original Acc. & AlexNet Patch & VGG Patch & ResNet18 Patch & ResNet101 Patch\\ \hline
BP & 100.00 & 100.00 & 100.00 & 100.00 & 100.00 \\ \hline
SIFT+FV & 87.00 & 81.00 & 78.00 & 84.00 & 81.00 \\ \hline
AlexNet & 94.00 & 24.00 & 85.00 & 86.00 & 77.00 \\ \hline
VGG & 95.00 & 41.00 & 19.00 & 48.00 & 23.00 \\ \hline
ResNet18  & 95.00 & 22.00 & 39.00 & 0.00 & 9.00 \\ \hline
ResNet101  & 96.00 & 43.00 & 41.00 & 35.00 & 22.00 \\ \hline
\hline
\textbf{Drawings} & Original Acc. & AlexNet Patch & VGG Patch & ResNet18 Patch & ResNet101 Patch\\ \hline
BP & 91.00 & 91.00 & 91.00 & 91.00 & 91.00 \\ \hline
SIFT+FV & 72.00 & 67.00 & 68.00 & 69.00 & 67.00 \\ \hline
AlexNet & 94.00 & 30.00 & 85.00 & 80.00 & 73.00 \\ \hline
VGG & 98.00 & 81.00 & 69.00 & 74.00 & 62.00 \\ \hline
ResNet18  & 96.00 & 82.00 & 91.00 & 66.00 & 79.00 \\ \hline
ResNet101  & 99.00 & 88.00 & 90.00 & 85.00 & 75.00 \\ \hline
\hline
\textbf{Sculpture} & Original Acc. & AlexNet Patch & VGG Patch & ResNet18 Patch & ResNet101 Patch\\ \hline
BP & 85.00 & 85.00 & 85.00 & 85.00 & 85.00 \\ \hline
SIFT+FV & 95.00 & 92.00 & 94.00 & 94.00 & 95.00 \\ \hline
AlexNet & 97.00 & 32.00 & 92.00 & 89.00 & 86.00 \\ \hline
VGG & 97.00 & 93.00 & 72.00 & 85.00 & 85.00 \\ \hline
ResNet18  & 95.00 & 92.00 & 86.00 & 66.00 & 89.00 \\ \hline
ResNet101  & 94.00 & 87.00 & 89.00 & 86.00 & 87.00 \\ \hline
\hline
\textbf{Engraving BW} & Original Acc. & AlexNet Patch & VGG Patch & ResNet18 Patch & ResNet101 Patch\\ \hline
BP & 90.00 & 90.00 & 91.00 & 91.00 & 91.00 \\ \hline
SIFT+FV & 91.00 & 94.00 & 93.00 & 95.00 & 92.00 \\ \hline
AlexNet & 100.00 & 99.00 & 100.00 & 100.00 & 100.00 \\ \hline
VGG & 100.00 & 99.00 & 83.00 & 96.00 & 97.00 \\ \hline
ResNet18  & 100.00 & 100.00 & 96.00 & 71.00 & 96.00 \\ \hline
ResNet101  & 100.00 & 100.00 & 100.00 & 100.00 & 100.00 \\ \hline
\hline
\textbf{Engraving Color} & Original Acc. & AlexNet Patch & VGG Patch & ResNet18 Patch & ResNet101 Patch\\ \hline
BP & 93.00 & 92.00 & 93.00 & 92.00 & 92.00 \\ \hline
SIFT+FV & 94.00 & 94.00 & 96.00 & 95.00 & 95.00 \\ \hline
AlexNet & 97.00 & 67.00 & 98.00 & 95.00 & 93.00 \\ \hline
VGG & 100.00 & 100.00 & 99.00 & 100.00 & 100.00 \\ \hline
ResNet18  & 98.00 & 99.00 & 100.00 & 98.00 & 99.00 \\ \hline
ResNet101  & 100.00 & 100.00 & 100.00 & 100.00 & 100.00 \\ \hline
\end{tabular}
}
 \end{center}
\end{table}

\subsubsection{Statistical Analysis of Robustness}

In the last section, we see that differences among experiments seem striking, particularly when images suffer a subtle perturbation. Nevertheless, statistical analysis allows us to be more confident regarding the robustness of each method's predictions. Nowadays, the nonparametric statistical analysis is bringing researchers' attention to measure the performance through a rigorous comparison among algorithms, considering independence, normality, and homoscedasticity \cite{Derrac2011APT,Garca2009ASO}. Such procedures perform both pairwise and multiple comparisons for multiple-problem analysis. In our case, we apply pairwise statistical procedures to perform individual comparisons between each method's predictions' confidence from clean and attacked images based on the statistical procedure described in \cite{Vega2020TimeAI}.

When the designed algorithms' results for the same problem achieved the conditions expressed before, the most common test is the ANOVA. In case that the distributions are not normal, we must use a nonparametric test like Kruskal-Wallis. If the distributions are normal but do not achieve the property of homoscedasticity, the analysis required is the Welch test. The statistical tests enable comparisons of the sample distributions, attending to the required conditions, and applying a suitable assessment a posteriori to contrast the results. As a result, we have first studied data normality (Lilliefors, Kolmogorov-Smirnov) and homoscedasticity (Levene test); then, according to the results, we have applied the appropriate statistical test (Kruskal-Wallis, Welch, Anova) to determine if the differences are significant, using a p-value $ < 0.05$. Therefore, if the predictions' confidence is statistically different, it will illustrate the rejection of the null hypothesis $Ho$. If the statistical analysis accepts $Ho$, it will define that the predictions' confidence from the pair of clean and perturbed images is not significantly different; hence we can conclude that the method is robust to the AEs.

The statistical analysis from Tables~\ref{tabrobust}-\ref{tabrobust2} shows that the predictions' confidence from BP is not significantly different in every experiment of the test dataset using FGSM. That means that the confidence is not affected by the subtle perturbations added to the images. The majority of p-values from SIFT+FV demonstrate to be not significantly different between the predictions' confidences.  Nonetheless, the analysis from all DCNN architectures showed that, in most cases, the rejection of the null hypothesis $Ho$. The rejection illustrates the damage of the AEs to the DCNN's predictions' confidence by making them statistically different.

\begin{table}[H]
\caption{Results from the statistical tests applied to each method's predictions' confidence from clean and attacked images using test datasets and AEs from AlexNet and VGG. Each value represents the corresponding p-value from the statistical test.}
\label{tabrobust}
\begin{center}
\resizebox{\textwidth}{!}
{
\begin{tabular}{|l|l|l|l|l|l|l|l|l|l|l|}
\hline
\multicolumn{11}{|c|}{Iconography}  \\ \hline
& \multicolumn{5}{c|}{AlexNet} & \multicolumn{5}{c|}{VGG}\\ \hline
Testing Vs. & $\epsilon2$ & $\epsilon4$ & $\epsilon8$ & $\epsilon16$ & $\epsilon32$ & $\epsilon2$ & $\epsilon4$ & $\epsilon8$ & $\epsilon16$ & $\epsilon32$  \\ 
\hline 
BP & 0.99743 & 0.99889 & 0.99599 & 0.9958 & 0.9958 & 0.99828 & 0.99634 & 0.99517 & 0.9937 & 0.99371 \\ \hline
SIFT+FV & 0.95119 & 0.95118 & 0.80859 & 0.21374 & 0.00012958 & 0.93022 & 0.93022 & 0.73482 & 0.10472 & 9.0663e-06 \\ \hline
AlexNet & 1.7665e-12 & 1.7665e-12 & 1.8922e-54 & 5.2622e-73 & 1.5556e-74 & 0.99483 & 0.99483 & 0.99702 & 0.84401 & 0.83964 \\ \hline
VGG & 0.58235 & 0.58245 & 0.0087004 & 1.884e-09 & 3.1824e-32 & 3.2201e-30 & 3.2201e-30 & 3.1832e-64 & 1.3611e-174 & 7.6365e-176 \\ \hline
ResNet18 & 0.63062 & 0.63063 & 0.024056 & 8.0612e-06 & 3.3087e-17 & 0.52182 & 0.52182 & 0.00035193 & 3.2009e-10 & 2.4186e-22 \\ \hline
ResNet101 & 0.6259 & 0.62599 & 0.054054 & 5.9926e-05 & 2.5752e-11 & 0.54501 & 0.54501 & 0.0017698 & 1.0084e-06 & 2.232e-12 \\ \hline
\multicolumn{11}{|c|}{Painting}   \\ \hline
& \multicolumn{5}{c|}{AlexNet} & \multicolumn{5}{c|}{VGG}\\ \hline
Testing Vs. & $\epsilon2$ & $\epsilon4$ & $\epsilon8$ & $\epsilon16$ & $\epsilon32$ & $\epsilon2$ & $\epsilon4$ & $\epsilon8$ & $\epsilon16$ & $\epsilon32$  \\ 
\hline 
BP & 0.9136 & 0.89874 & 0.89634 & 0.85943 & 0.12557 & 0.99068 & 0.99108 & 0.78341 & 0.75145 & 0.59411 \\ \hline
SIFT+FV & 0.21866 & 0.21866 & 0.00012449 & 1.6086e-24 & 1.65e-86 & 0.16502 & 0.16502 & 3.9752e-06 & 1.516e-31 & 3.6495e-98 \\ \hline
AlexNet & 3.4962e-105 & 3.4962e-105 & 0 & 0 & 0 & 0.98106 & 0.98106 & 0.90152 & 0.47591 & 0.050199 \\ \hline
VGG & 0.64622 & 0.64621 & 9.6649e-14 & 2.4111e-137 & 0 & 0 & 0 & 0 & 0 & 0 \\ \hline
ResNet18 & 0.9711 & 0.97111 & 0.92035 & 0.13873 & 3.8914e-111 & 0.37065 & 0.37065 & 2.124e-25 & 1.5714e-103 & 1.1938e-291 \\ \hline
ResNet101 & 0.90558 & 0.90557 & 0.37347 & 3.1411e-66 & 4.1528e-240 & 0.35338 & 0.35338 & 3.3466e-53 & 1.3622e-216 & 0 \\ \hline
\multicolumn{11}{|c|}{Painting Landscapes}   \\ \hline
& \multicolumn{5}{c|}{AlexNet} & \multicolumn{5}{c|}{VGG}\\ \hline
Testing Vs. & $\epsilon2$ & $\epsilon4$ & $\epsilon8$ & $\epsilon16$ & $\epsilon32$ & $\epsilon2$ & $\epsilon4$ & $\epsilon8$ & $\epsilon16$ & $\epsilon32$  \\ 
\hline 
BP & 1 & 1 & 1 & 1 & 1 & 1 & 1 & 1 & 1 & 1 \\ \hline
SIFT+FV & 0.79505 & 0.79505 & 0.30022 & 0.0024393 & 3.3167e-09 & 0.79599 & 0.79599 & 0.3353 & 0.0019852 & 3.6761e-09 \\ \hline
AlexNet & 7.9291e-10 & 7.9291e-10 & 1.4342e-29 & 5.265e-33 & 8.3707e-33 & 0.9905 & 0.9905 & 0.95413 & 0.95212 & 0.70978 \\ \hline
VGG & 0.87445 & 0.87447 & 0.38584 & 3.5473e-12 & 7.3903e-34 & 5.0409e-29 & 5.0409e-29 & 5.5977e-37 & 3.0354e-37 & 1.0586e-38 \\ \hline
ResNet18 & 0.89967 & 0.89966 & 0.7306 & 0.30211 & 1.4175e-05 & 0.68713 & 0.68713 & 0.030198 & 1.4375e-05 & 7.5887e-15 \\ \hline
ResNet101 & 0.9671 & 0.96696 & 0.74545 & 5.8498e-08 & 2.5216e-22 & 0.87901 & 0.87901 & 0.33332 & 2.368e-17 & 3.0898e-30 \\ \hline
\multicolumn{11}{|c|}{Drawings}   \\ \hline
& \multicolumn{5}{c|}{AlexNet} & \multicolumn{5}{c|}{VGG}\\ \hline
Testing Vs. & $\epsilon2$ & $\epsilon4$ & $\epsilon8$ & $\epsilon16$ & $\epsilon32$ & $\epsilon2$ & $\epsilon4$ & $\epsilon8$ & $\epsilon16$ & $\epsilon32$  \\ 
\hline 
BP & 0.98405 & 0.98405 & 0.97876 & 0.95426 & 0.97483 & 0.98745 & 0.98745 & 0.99322 & 0.98889 & 0.97186 \\ \hline
SIFT+FV & 0.67854 & 0.67854 & 0.43269 & 0.04983 & 2.6184e-06 & 0.69315 & 0.69315 & 0.46697 & 0.064349 & 1.0753e-05 \\ \hline
AlexNet & 1.5308e-05 & 1.5308e-05 & 3.0146e-26 & 5.9791e-41 & 6.2069e-47 & 0.92991 & 0.92989 & 0.83881 & 0.84904 & 0.0052409 \\ \hline
VGG & 0.87066 & 0.87066 & 0.058863 & 9.9068e-08 & 3.7992e-30 & 1.5939e-20 & 1.5939e-20 & 1.7229e-53 & 3.4676e-58 & 6.0516e-61 \\ \hline
ResNet18 & 0.90781 & 0.90781 & 0.50143 & 0.0023366 & 5.16e-16 & 0.59103 & 0.59103 & 0.0056782 & 6.2775e-08 & 9.5421e-26 \\ \hline
ResNet101 & 0.97912 & 0.97912 & 0.55583 & 3.2526e-06 & 1.313e-22 & 0.79099 & 0.79099 & 0.00070622 & 4.8929e-12 & 1.5062e-33 \\ \hline
\multicolumn{11}{|c|}{Sculpture}   \\ \hline
& \multicolumn{5}{c|}{AlexNet} & \multicolumn{5}{c|}{VGG}\\ \hline
Testing Vs. & $\epsilon2$ & $\epsilon4$ & $\epsilon8$ & $\epsilon16$ & $\epsilon32$ & $\epsilon2$ & $\epsilon4$ & $\epsilon8$ & $\epsilon16$ & $\epsilon32$  \\ 
\hline 
BP & 0.99772 & 0.99772 & 0.99742 & 0.99748 & 0.99701 & 0.99736 & 0.99736 & 0.99717 & 0.99702 & 0.99753 \\ \hline
SIFT+FV & 0.89198 & 0.89198 & 0.4632 & 0.1714 & 0.086323 & 0.93114 & 0.93114 & 0.5201 & 0.20871 & 0.14017 \\ \hline
AlexNet & 0.00013513 & 0.00013513 & 2.0152e-20 & 8.8035e-31 & 8.2946e-34 & 0.89484 & 0.89484 & 0.54423 & 0.00026161 & 1.6579e-12 \\ \hline
VGG & 0.75594 & 0.75594 & 0.0082156 & 5.2489e-08 & 1.6714e-21 & 6.7774e-13 & 6.7774e-13 & 1.0801e-30 & 7.3611e-34 & 2.2997e-36 \\ \hline
ResNet18 & 0.76905 & 0.76905 & 0.053247 & 2.3723e-06 & 7.1788e-18 & 0.6501 & 0.6501 & 0.0032158 & 1.3364e-08 & 4.5108e-21 \\ \hline
ResNet101 & 0.83153 & 0.83153 & 0.018017 & 6.1378e-07 & 2.1775e-14 & 0.73046 & 0.73046 & 0.0025386 & 2.3318e-07 & 3.531e-15 \\ \hline
\multicolumn{11}{|c|}{Engraving BW}   \\ \hline
& \multicolumn{5}{c|}{AlexNet} & \multicolumn{5}{c|}{VGG}\\ \hline
Testing Vs. & $\epsilon2$ & $\epsilon4$ & $\epsilon8$ & $\epsilon16$ & $\epsilon32$ & $\epsilon2$ & $\epsilon4$ & $\epsilon8$ & $\epsilon16$ & $\epsilon32$  \\ 
\hline 
BP & 0.1314 & 0.1314 & 0.75089 & 0.7213 & 0.76281 & 0.1184 & 0.1184 & 0.69932 & 0.72718 & 0.75278 \\ \hline
SIFT+FV & 0.89606 & 0.89606 & 0.9213 & 0.77775 & 0.01329 & 0.87447 & 0.87447 & 0.85628 & 0.96241 & 0.086893 \\ \hline
AlexNet & 5.358e-33 & 5.358e-33 & 2.205e-116 & 3.5771e-171 & 2.9344e-187 & 0.61108 & 0.61108 & 0.79384 & 0.66623 & 1.2529e-07 \\ \hline
VGG & 0.94146 & 0.94146 & 0.62982 & 0.019374 & 0.092515 & 1.7458e-85 & 1.7458e-85 & 2.3367e-202 & 6.0154e-219 & 3.4579e-219 \\ \hline
ResNet18 & 0.35122 & 0.35122 & 0.020752 & 6.4226e-13 & 4.1326e-56 & 0.11378 & 0.11378 & 2.9188e-05 & 5.2976e-22 & 1.1364e-73 \\ \hline
ResNet101 & 0.14262 & 0.14262 & 0.70567 & 0.0010867 & 5.2621e-31 & 0.4537 & 0.4537 & 0.19041 & 4.4468e-06 & 1.2934e-39 \\ \hline
\multicolumn{11}{|c|}{Engraving Color}  \\ \hline
& \multicolumn{5}{c|}{AlexNet} & \multicolumn{5}{c|}{VGG}\\ \hline
Testing Vs. & $\epsilon2$ & $\epsilon4$ & $\epsilon8$ & $\epsilon16$ & $\epsilon32$ & $\epsilon2$ & $\epsilon4$ & $\epsilon8$ & $\epsilon16$ & $\epsilon32$  \\ 
\hline 
BP & 0.89591 & 0.89591 & 0.94823 & 0.93453 & 0.98186 & 0.87265 & 0.87265 & 0.92126 & 0.93354 & 0.99179 \\ \hline
SIFT+FV & 0.83715 & 0.83715 & 0.7306 & 0.52109 & 0.95255 & 0.78356 & 0.78356 & 0.78447 & 1 & 0.82826 \\ \hline
AlexNet & 1.6503e-133 & 1.6479e-133 & 4.924e-273 & 2.106e-290 & 1.6302e-274 & 0.56852 & 0.56852 & 0.36771 & 0.041684 & 1.0253e-05 \\ \hline
VGG & 3.66e-32 & 3.66e-32 & 1.8892e-35 & 1.7963e-34 & 1.0132e-18 & 0 & 0 & 0 & 0 & 0 \\ \hline
ResNet18 & 1.105e-08 & 1.105e-08 & 1.8668e-09 & 4.1096e-08 & 6.1218e-06 & 1.6772e-10 & 1.6772e-10 & 1.7619e-14 & 1.0486e-16 & 1.5263e-15 \\ \hline
ResNet101 & 5.1941e-16 & 5.1941e-16 & 1.1333e-18 & 6.1089e-25 & 5.416e-42 & 8.2181e-23 & 8.2181e-23 & 1.968e-41 & 3.066e-70 & 3.3115e-107 \\ 
\hline
\end{tabular}}
\end{center}
\end{table}

\begin{table}[H]
\caption{Results from the statistical tests applied to each method's predictions' confidence from clean and attacked images using test datasets and AEs from ResNet and ResNet101. Each value represents the corresponding p-value from the statistical test.}
\label{tabrobust2}
\begin{center}
\resizebox{\textwidth}{!}
{
\begin{tabular}{|l|l|l|l|l|l|l|l|l|l|l|}
\hline
\multicolumn{11}{|c|}{Iconography}  \\ \hline
& \multicolumn{5}{c|}{ResNet} & \multicolumn{5}{c|}{ResNet101}\\ \hline
Testing Vs. & $\epsilon2$ & $\epsilon4$ & $\epsilon8$ & $\epsilon16$ & $\epsilon32$ & $\epsilon2$ & $\epsilon4$ & $\epsilon8$ & $\epsilon16$ & $\epsilon32$  \\ 
\hline 
BP & 0.99767 & 0.99672 & 0.99739 & 0.99585 & 0.99585 & 0.99992 & 0.99593 & 0.99825 & 0.99589 & 0.99589 \\ \hline 
SIFT+FV & 0.95457 & 0.95457 & 0.78501 & 0.098796 & 1.296e-05 & 0.9245 & 0.92449 & 0.71244 & 0.06041 & 2.3283e-06 \\ \hline 
AlexNet & 0.99776 & 0.99776 & 0.96299 & 0.74961 & 0.18328 & 0.9903 & 0.9903 & 0.98065 & 0.79478 & 0.62045 \\ \hline 
VGG & 0.45019 & 0.45018 & 1.3824e-05 & 7.6022e-17 & 2.9553e-39 & 0.60766 & 0.60768 & 0.0020685 & 3.2656e-09 & 6.6815e-28 \\ \hline 
ResNet18 & 2.8465e-35 & 2.8465e-35 & 1.2532e-66 & 1.0059e-161 & 1.2666e-66 & 0.48618 & 0.48626 & 0.0005933 & 1.195e-08 & 9.7675e-18 \\ \hline 
ResNet101 & 0.36979 & 0.36979 & 1.161e-05 & 8.3498e-14 & 2.6481e-22 & 5.2645e-31 & 5.2645e-31 & 1.3736e-58 & 7.7789e-61 & 7.2518e-59 \\ \hline
\multicolumn{11}{|c|}{Painting}   \\ \hline
& \multicolumn{5}{c|}{ResNet} & \multicolumn{5}{c|}{ResNet101}\\ \hline
Testing Vs. & $\epsilon2$ & $\epsilon4$ & $\epsilon8$ & $\epsilon16$ & $\epsilon32$ & $\epsilon2$ & $\epsilon4$ & $\epsilon8$ & $\epsilon16$ & $\epsilon32$  \\ 
\hline 
BP & 0.84987 & 0.78268 & 0.54929 & 0.99293 & 0.9937 & 0.73759 & 0.94754 & 0.87769 & 0.88685 & 0.83682 \\ \hline 
SIFT+FV & 0.15684 & 0.15685 & 1.1729e-06 & 9.2235e-34 & 4.2265e-102 & 0.1448 & 0.1448 & 4.1104e-07 & 1.6025e-34 & 6.9011e-99 \\ \hline 
AlexNet & 0.97537 & 0.97538 & 0.83884 & 0.0002221 & 0.056476 & 0.98499 & 0.98501 & 0.8621 & 0.39927 & 0.11822 \\ \hline 
VGG & 1.5516e-06 & 1.5524e-06 & 1.4289e-58 & 1.8098e-237 & 0 & 8.3796e-06 & 8.3735e-06 & 5.48e-50 & 1.1918e-222 & 0 \\ \hline 
ResNet18 & 0 & 0 & 0 & 0 & 0 & 9.7705e-05 & 9.7674e-05 & 3.6504e-28 & 2.1311e-107 & 9.0869e-291 \\ \hline 
ResNet101 & 0.20263 & 0.20266 & 7.6609e-70 & 1.041e-267 & 0 & 0 & 0 & 0 & 0 & 0 \\  \hline
\multicolumn{11}{|c|}{Painting Landscapes}   \\ \hline
& \multicolumn{5}{c|}{ResNet} & \multicolumn{5}{c|}{ResNet101}\\ \hline
Testing Vs. & $\epsilon2$ & $\epsilon4$ & $\epsilon8$ & $\epsilon16$ & $\epsilon32$ & $\epsilon2$ & $\epsilon4$ & $\epsilon8$ & $\epsilon16$ & $\epsilon32$  \\ 
\hline 
BP & 1 & 1 & 1 & 1 & 1 & 1 & 1 & 1 & 1 & 1 \\ \hline 
SIFT+FV & 0.81073 & 0.81072 & 0.31041 & 0.0013292 & 9.4406e-10 & 0.79544 & 0.79544 & 0.27643 & 0.0011679 & 3.2551e-09 \\ \hline 
AlexNet & 0.97698 & 0.97699 & 0.93679 & 0.92671 & 0.66599 & 0.99458 & 0.9946 & 0.97321 & 0.97355 & 0.81807 \\ \hline 
VGG & 0.71643 & 0.71643 & 4.7994e-05 & 1.6994e-17 & 4.5878e-38 & 0.73709 & 0.73709 & 0.00021444 & 4.9845e-16 & 8.8656e-38 \\ \hline 
ResNet18 & 1.3222e-24 & 1.3435e-24 & 1.8235e-34 & 5.3646e-34 & 2.2238e-35 & 0.60616 & 0.60616 & 0.021885 & 2.1956e-05 & 1.1705e-16 \\ \hline 
ResNet101 & 0.83154 & 0.83154 & 1.4205e-06 & 5.6908e-22 & 5.671e-33 & 8.0509e-29 & 8.0509e-29 & 7.9778e-37 & 9.0441e-38 & 5.7158e-41 \\  \hline
\multicolumn{11}{|c|}{Drawings}   \\ \hline
& \multicolumn{5}{c|}{ResNet} & \multicolumn{5}{c|}{ResNet101}\\ \hline
Testing Vs. & $\epsilon2$ & $\epsilon4$ & $\epsilon8$ & $\epsilon16$ & $\epsilon32$ & $\epsilon2$ & $\epsilon4$ & $\epsilon8$ & $\epsilon16$ & $\epsilon32$  \\ 
\hline 
BP & 0.98646 & 0.98646 & 0.98853 & 0.96403 & 0.97252 & 0.99539 & 0.99539 & 0.99544 & 0.98031 & 0.98827 \\ \hline 
SIFT+FV & 0.69545 & 0.69544 & 0.45485 & 0.052022 & 4.3891e-06 & 0.68791 & 0.68791 & 0.43034 & 0.039834 & 2.482e-06 \\ \hline 
AlexNet & 0.91835 & 0.91835 & 0.86539 & 0.77007 & 0.0013548 & 0.9221 & 0.92208 & 0.86883 & 0.94347 & 0.29188 \\ \hline 
VGG & 0.61463 & 0.61463 & 0.00064209 & 2.228e-13 & 1.8494e-35 & 0.5806 & 0.58052 & 0.0061323 & 1.4804e-10 & 5.7541e-34 \\ \hline 
ResNet18 & 1.7101e-24 & 1.7101e-24 & 1.9766e-51 & 1.4251e-54 & 1.2956e-56 & 0.50031 & 0.50028 & 0.00714 & 1.1845e-06 & 2.6572e-20 \\ \hline 
ResNet101 & 0.59366 & 0.59366 & 0.0001009 & 4.6116e-12 & 1.5527e-34 & 4.7426e-30 & 4.7426e-30 & 3.8037e-54 & 1.3584e-57 & 3.9703e-60 \\ \hline
\multicolumn{11}{|c|}{Sculpture}   \\ \hline
& \multicolumn{5}{c|}{ResNet} & \multicolumn{5}{c|}{ResNet101}\\ \hline
Testing Vs. & $\epsilon2$ & $\epsilon4$ & $\epsilon8$ & $\epsilon16$ & $\epsilon32$ & $\epsilon2$ & $\epsilon4$ & $\epsilon8$ & $\epsilon16$ & $\epsilon32$  \\ 
\hline 
BP & 0.9975 & 0.9975 & 0.99702 & 0.99693 & 0.9969 & 0.99659 & 0.99659 & 0.99686 & 0.99662 & 0.99609 \\ \hline 
SIFT+FV & 0.91246 & 0.91245 & 0.45842 & 0.19418 & 0.23847 & 0.91303 & 0.91303 & 0.40899 & 0.11917 & 0.0022884 \\ \hline 
AlexNet & 0.89915 & 0.89911 & 0.57069 & 0.00029538 & 2.4554e-12 & 0.92807 & 0.92807 & 0.63864 & 0.0012897 & 1.0607e-10 \\ \hline 
VGG & 0.71495 & 0.71495 & 0.00025963 & 4.4615e-10 & 1.6644e-23 & 0.83568 & 0.83568 & 0.0062375 & 1.065e-06 & 8.4707e-20 \\ \hline 
ResNet18 & 5.4653e-13 & 5.4653e-13 & 1.885e-26 & 2.7286e-29 & 2.1562e-32 & 0.8016 & 0.8016 & 0.01385 & 2.2828e-06 & 1.1366e-17 \\ \hline 
ResNet101 & 0.63809 & 0.63809 & 0.00097214 & 2.3315e-09 & 1.5158e-17 & 7.4724e-11 & 7.4724e-11 & 1.7682e-19 & 8.4807e-21 & 1.7814e-25 \\ \hline
\multicolumn{11}{|c|}{Engraving BW}   \\ \hline
& \multicolumn{5}{c|}{ResNet} & \multicolumn{5}{c|}{ResNet101}\\ \hline
Testing Vs. & $\epsilon2$ & $\epsilon4$ & $\epsilon8$ & $\epsilon16$ & $\epsilon32$ & $\epsilon2$ & $\epsilon4$ & $\epsilon8$ & $\epsilon16$ & $\epsilon32$  \\ 
\hline 
BP & 0.62139 & 0.62139 & 0.76279 & 0.72754 & 0.73081 & 0.23525 & 0.23525 & 0.73697 & 0.75743 & 0.76744 \\ \hline 
SIFT+FV & 0.88575 & 0.88575 & 0.8536 & 0.95828 & 0.14382 & 0.86903 & 0.86904 & 0.88072 & 0.92791 & 0.065097 \\ \hline 
AlexNet & 0.88255 & 0.88255 & 0.82907 & 0.12103 & 5.201e-14 & 0.63177 & 0.63178 & 0.85028 & 0.51736 & 0.026006 \\ \hline 
VGG & 0.3549 & 0.3549 & 0.011117 & 1.113e-11 & 1.2312e-46 & 0.61302 & 0.61311 & 0.14793 & 1.5966e-08 & 1.9818e-32 \\ \hline 
ResNet18 & 2.0619e-81 & 2.0619e-81 & 8.5802e-191 & 2.2357e-211 & 4.2645e-218 & 0.10882 & 0.10882 & 0.0001122 & 9.641e-23 & 5.6127e-80 \\ \hline 
ResNet101 & 0.5792 & 0.5792 & 0.0092807 & 5.08e-14 & 5.8058e-77 & 4.4317e-56 & 4.4223e-56 & 1.1988e-119 & 1.3737e-152 & 3.8698e-193 \\  \hline
Testing Vs. & $\epsilon2$ & $\epsilon4$ & $\epsilon8$ & $\epsilon16$ & $\epsilon32$ & $\epsilon2$ & $\epsilon4$ & $\epsilon8$ & $\epsilon16$ & $\epsilon32$  \\ 
\hline 
BP & 0.96553 & 0.96553 & 0.93342 & 0.99733 & 0.99721 & 0.96275 & 0.96275 & 0.9169 & 0.98099 & 0.83034 \\ \hline 
SIFT+FV & 0.89023 & 0.89023 & 0.67872 & 0.66967 & 0.8903 & 0.89075 & 0.89075 & 0.72948 & 0.51895 & 0.8264 \\ \hline 
AlexNet & 0.61061 & 0.61064 & 0.4212 & 0.095421 & 0.0014581 & 0.56698 & 0.56703 & 0.33301 & 0.017158 & 1.6441e-06 \\ \hline 
VGG & 8.8075e-36 & 8.8007e-36 & 1.3014e-44 & 7.2256e-54 & 7.1826e-59 & 1.3175e-40 & 1.323e-40 & 6.4181e-58 & 1.2544e-79 & 1.8917e-105 \\ \hline 
ResNet18 & 3.076e-307 & 3.076e-307 & 0 & 0 & 0 & 3.907e-11 & 3.907e-11 & 5.3445e-16 & 1.6623e-17 & 1.5303e-14 \\ \hline 
ResNet101 & 4.2e-20 & 4.1976e-20 & 4.9361e-32 & 4.2834e-52 & 6.6939e-89 & 2.6025e-310 & 2.6206e-310 & 0 & 0 & 0 \\  
\hline
\end{tabular}}
\end{center}
\end{table}

The same behavior is seen in the statistical analysis from Table~\ref{tabrobust3}, which shows the method's predictions' confidence to the adversarial patch. The study showed the same rejection of the null hypothesis $Ho$ from all DCNN architectures in a significant part of the experiments for all classes. Conversely, BP accepted the null hypothesis $Ho$ in every experiment. SIFT+FV showed similar behavior to BP, but the sculpture class and the VGG patch obtained significantly different predictions' confidence.

\begin{table}
\caption{P-value obtained using the adversarial patch. Each column presents the score obtained for the 100 images per pair (Clean and AE) using the adversarial patch. }
\label{tabrobust3}
\begin{center}
\resizebox{0.7\textwidth}{!}
{
\begin{tabular}{|l|l|l|l|l|}
\hline
\multicolumn{5}{|c|}{\textbf{Iconography}}\\
\hline
Testing Vs. & AlexNet Patch & VGG Patch & ResNet18 Patch & ResNet101 Patch\\ \hline
BP & 0.99608 & 0.99019 & 0.99812 & 0.99901 \\ \hline
SIFT+FV & 0.60131 & 0.89837 & 0.61682 & 0.97909 \\ \hline 
AlexNet & 7.1039e-20 & 0.79567 & 0.46915 & 0.56209 \\ \hline 
VGG & 0.0003743 & 1.3363e-23 & 9.9987e-07 & 1.0847e-07 \\ \hline 
ResNet18 & 2.5542e-05 & 0.00040313 & 2.092e-18 & 0.00029613 \\ \hline 
ResNet101 & 0.0010507 & 8.4448e-05 & 3.9575e-08 & 1.358e-12 \\ 
\hline
\multicolumn{5}{|c|}{\textbf{Painting}}\\
\hline
Testing Vs. & AlexNet Patch & VGG Patch & ResNet18 Patch & ResNet101 Patch\\ \hline
BP & 1 & 1 & 1 & 1 \\ \hline 
SIFT+FV & 0.54497 & 0.74837 & 0.55432 & 0.82595 \\ \hline 
AlexNet & 1.3914e-27 & 2.7344e-07 & 4.1336e-07 & 7.5744e-12 \\ \hline 
VGG & 4.5007e-16 & 3.0804e-22 & 5.2464e-14 & 3.4408e-20 \\ \hline 
ResNet18 & 1.2746e-19 & 1.1991e-15 & 2.269e-29 & 4.9202e-26 \\ \hline 
ResNet101 & 3.855e-20 & 1.9447e-20 & 5.876e-18 & 9.657e-26 \\ 
\hline
\multicolumn{5}{|c|}{\textbf{Painting Landscapes}}\\
\hline
Testing Vs. & AlexNet Patch & VGG Patch & ResNet18 Patch & ResNet101 Patch\\ \hline
BP & 1 & 1 & 1 & 1 \\ \hline 
SIFT+FV & 0.71106 & 0.35442 & 0.76126 & 0.32247 \\ \hline 
AlexNet & 2.3132e-27 & 1.4598e-08 & 1.3614e-06 & 4.8304e-13 \\ \hline 
VGG & 3.4926e-25 & 5.7961e-30 & 1.6104e-22 & 3.0764e-29 \\ \hline 
ResNet18 & 1.5553e-28 & 1.3091e-24 & 1.1945e-33 & 5.1204e-32 \\ \hline 
ResNet101 & 1.868e-27 & 5.565e-27 & 3.9203e-28 & 1.1313e-30 \\ 
\hline
\multicolumn{5}{|c|}{\textbf{Drawings}}\\
\hline
Testing Vs. & AlexNet Patch & VGG Patch & ResNet18 Patch & ResNet101 Patch\\ \hline
BP & 1 & 1 & 1 & 1 \\ \hline 
SIFT+FV & 0.49234 & 0.24792 & 0.40336 & 0.20665 \\ \hline 
AlexNet & 2.0788e-27 & 2.61e-05 & 2.1405e-08 & 7.0741e-12 \\ \hline 
VGG & 3.9051e-15 & 2.6781e-20 & 1.1136e-18 & 3.3552e-21 \\ \hline 
ResNet18 & 5.2286e-15 & 3.8956e-09 & 1.3091e-24 & 5.042e-17 \\ \hline 
ResNet101 & 3.8469e-17 & 9.1359e-12 & 5.7602e-14 & 1.0681e-22 \\ 
\hline
\multicolumn{5}{|c|}{\textbf{Sculpture}}\\
\hline
Testing Vs. & AlexNet Patch & VGG Patch & ResNet18 Patch & ResNet101 Patch\\ \hline
BP & 0.99544 & 0.99701 & 0.99823 & 0.99557 \\ \hline 
SIFT+FV & 0.093228 & 0.048075 & 0.14621 & 0.49341 \\ \hline 
AlexNet & 9.0533e-28 & 1.6758e-06 & 1.6128e-08 & 8.3985e-08 \\ \hline 
VGG & 6.2478e-06 & 4.5278e-19 & 1.1145e-10 & 3.8384e-09 \\ \hline 
ResNet18 & 9.011e-06 & 5.6851e-07 & 3.0015e-20 & 6.7596e-08 \\ \hline 
ResNet101 & 3.1949e-07 & 0.00045029 & 2.9946e-07 & 1.2838e-08 \\
\hline
\multicolumn{5}{|c|}{\textbf{Engraving BW}}\\
\hline
Testing Vs. & AlexNet Patch & VGG Patch & ResNet18 Patch & ResNet101 Patch\\ \hline
BP & 0.84032 & 0.9292 & 0.83695 & 0.79371 \\ \hline 
SIFT+FV & 0.17899 & 0.68646 & 0.63549 & 0.75774 \\ \hline 
AlexNet & 6.8616e-16 & 0.33203 & 0.13546 & 0.80088 \\ \hline 
VGG & 2.7464e-15 & 6.3779e-28 & 1.1879e-13 & 1.0857e-15 \\ \hline 
ResNet18 & 4.2622e-24 & 1.3125e-21 & 5.7495e-32 & 6.0573e-26 \\ \hline 
ResNet101 & 1.5343e-07 & 1.7745e-10 & 4.6019e-14 & 4.6746e-23 \\ 
\hline
\multicolumn{5}{|c|}{\textbf{Engraving Color}}\\
\hline
Testing Vs. & AlexNet Patch & VGG Patch & ResNet18 Patch & ResNet101 Patch\\ \hline
BP & 0.81987 & 0.96904 & 0.82055 & 0.81363 \\ \hline 
SIFT+FV & 0.51466 & 1 & 1 & 1 \\ \hline 
AlexNet & 4.7445e-15 & 0.42195 & 0.6454 & 0.32844 \\ \hline 
VGG & 0.00057199 & 5.3585e-08 & 0.11452 & 0.000143 \\ \hline 
ResNet18 & 0.00012499 & 0.01838 & 0.00042032 & 0.31804 \\ \hline 
ResNet101 & 0.33861 & 0.29463 & 1.2204e-05 & 8.898e-10 \\ \hline
\end{tabular}
}
 \end{center}
\end{table}

\section{Conclusion}
\label{sec:Conclusion}

Robustness against AA must be the primary concern when it is developing an automatic recognition system. So, from now on, a classifier's performance should not be focused only on accuracy but also on robustness to AAs. In this work, we present a comparative study for AMC subject to AA. We compare several methods to analyze the performance and their reliability to predict a class using adversarial perturbations. We selected six models using three of the main approaches for image classification: 1) handcrafted features approach (SIFT+FV), 2) deep genetic programming approach (BP), and 3) DCNN approach (AlexNet, VGG, ResNet18, and ResNet101). The comparative study consists of analyzing three different attacks. Firstly, the direct threat's impact and transferability considering the white box untargeted attack--FGSM. This perturbation adds a subtle texture to the artwork, which can cause a misleading prediction. Secondly, find a set of localization and pixel values to modify the artwork to fool the classifier using a black box untargeted attack--multiple pixel attack. Finally, apply precomputed patches--adversarial patch--robust to transformations located randomly in the artwork to predict a targeted class.

In this sense, this study has demonstrated that AA is a severe threat to the performance of DCNN. Using FGSM showed that if the attacker knows the model, it can make the DCNN decrease its performance up to less than 20\% of its original score. Additionally, we proved the transferability effect between DCNN models, which is not severe for the binary classification, but it can reduce up to 20\% of the performance. On the other hand, SIFT+FV also was affected by some of the classes but by a minor amount. However, the added texture caused by the FGSM lead to a decrease in its performance in a significant manner when the algorithm was tested, having encouraging results but not suitable to compete with DCNNs in the testing phase. Finally, BP exhibits comparable performance (efficiency) to DCNN in both validation and testing phases. It has an almost imperceptible variant on its accuracy to these perturbations proving no direct transferability from other models. Figures ~\ref{fig:dimensionResnet}-\ref{fig:dimensionOne} can be observed the output of each stage of BP from clean and AEs with almost no variation on its outcomes.

The study about one pixel attack confirms this type of attack's poor design due to a minimal scenario contrived with an input image of size $32 \times 32$ pixels. We conclude that it is challenging to apply multiple pixel attacks on real-world conditions. On the one hand, when we extend to multiple pixels, the perturbation loses the attack's intention of being imperceptible to human vision, not to mention the massive amount of processing time. On the other hand, BP shows it challenging to fail these attacks even by increasing five times the number of pixels per AA compared to SIFT+FV and DCNN models, which were successfully fooled. Finally, the adversarial patch showed that a precomputed perturbation positioned in a random location and orientation in the artwork could fool DCNN models with excellent transferability between them; meanwhile, BP and SIFT+FV remain in their original score. It is remarkable the BP robustness to the multiple pixel attack and the adversarial patch. However, these two attacks are harsh perturbations and BP remained steady in its performance, leading to the reliability of BP's predictions in no human supervision cases.

The statistical analysis from the predictions' confidence supports the study of robustness by illustrating the change in the posterior probability complementing the results from the accuracy's standpoint. In this manner, BP demonstrated to have not significantly different predictions' confidence compare to DCNN models, which showed in most cases the rejection of the null hypothesis $Ho$. Conversely, SIFT+FV obtained good results, with most of the test scoring a not a significant difference in the predictions' confidence.

In conclusion, art media categorization is a complex problem in which it is difficult to outperform DCNN performance. Still, BP has comparable results and is robust to these adversarial attacks with no direct transferability of such perturbations to the model. On the other hand, SIFT+FV proves to be robust for a limited number of experiments with moderate results. So, BP arises as an alternative proposal of an art media classifier without the vulnerabilities of AA. Additionally, it takes advantage of the symbolic representations and incorporates rules from expert systems in a hierarchical structure to solve the AMC problem. Lastly, BP opens the possibility of being explainable on each of its stages, unlike DCNN, an important research area to know precisely the model's inner workings.

\end{document}